\documentclass[letterpaper]{article} 
\usepackage{aaai24}  
\usepackage{times}  
\usepackage{helvet}  
\usepackage{courier}  
\usepackage[hyphens]{url}  
\usepackage{graphicx} 
\usepackage{natbib}  
\usepackage{caption} 
\usepackage{algorithm}
\usepackage{algorithmic}

\usepackage{booktabs}
\usepackage{tabularx}
\usepackage{multirow}
\usepackage{makecell}

\usepackage{newfloat}
\usepackage{listings}
\DeclareCaptionStyle{ruled}{labelfont=normalfont,labelsep=colon,strut=off} 
\floatstyle{ruled}
\newfloat{listing}{tb}{lst}{}
\floatname{listing}{Listing}
\title{CORECODE: A Common Sense Annotated Dialogue Dataset with Benchmark Tasks for Chinese Large Language Models}
\author{
    Dan Shi\textsuperscript{\rm 1},
    Chaobin You\textsuperscript{\rm 1},
    Jiantao Huang\textsuperscript{\rm 2},
    Taihao Li\textsuperscript{\rm 2},
    Deyi Xiong\textsuperscript{\rm 1}\thanks{~Corresponding author}\\
}
\affiliations{
    \textsuperscript{\rm 1}College of Intelligence and Computing, Tianjin University, Tianjin, China\\
    \textsuperscript{\rm 2}Zhejiang Lab, Hangzhou, China

    \{shidan, chaobinyou, dyxiong\}@tju.edu.cn,
    \{jthuang, lith\}@zhejianglab.com
}

\usepackage{bibentry}

\begin{document}

\maketitle

\begin{abstract}
As an indispensable ingredient of intelligence, commonsense reasoning is crucial for large language models (LLMs) in real-world scenarios. In this paper, we propose CORECODE, a dataset that contains abundant commonsense knowledge manually annotated on dyadic dialogues, to evaluate the \textbf{co}mmonsense \textbf{re}asoning and commonsense \textbf{co}nflict \textbf{de}tection capabilities of Chinese LLMs. We categorize commonsense knowledge in everyday conversations into three dimensions: entity, event, and social interaction. For easy and consistent annotation, we standardize the form of commonsense knowledge annotation in open-domain dialogues as ``domain: slot = value''. A total of 9 domains and 37 slots are defined to capture diverse commonsense knowledge. With these pre-defined domains and slots, we collect 76,787 commonsense knowledge annotations from 19,700 dialogues through crowdsourcing. To evaluate and enhance the commonsense reasoning capability for LLMs on the curated dataset, we establish a series of dialogue-level reasoning and detection tasks, including commonsense knowledge filling, commonsense knowledge generation, commonsense conflict phrase detection, domain identification, slot identification, and event causal inference. A wide variety of existing open-source Chinese LLMs are evaluated with these tasks on our dataset. Experimental results demonstrate that these models are not competent to predict CORECODE's plentiful reasoning content, and even ChatGPT could only achieve 0.275 and 0.084 accuracy on the domain identification and slot identification tasks under the zero-shot setting. We release the data and codes of CORECODE at https://github.com/danshi777/CORECODE to promote commonsense reasoning evaluation and study of LLMs in the context of daily conversations. 
\end{abstract}

\section{Introduction}

Commonsense reasoning is a crucial component of intelligence\cite{liu2004conceptnet, Cambria2011IsanetteAC, Storks2019RecentAI}, which involves the ability to make logical deductions, infer implicit information and apply background knowledge to solve problems as well as understand the world. In recent years, exploring and improving the ability of NLP models for the acquisition and application of commonsense knowledge has been attracting growing interest, leading to extensive research in this field \cite{lin-etal-2019-kagnet, bauer_commonsense_2018, lv_graph-based_2020, wang_connecting_2020, liu_kg-bart_2021, jiang_im_2021, liu_generated_2022}.

It is widely acknowledged that LLMs, trained on a huge amount of data, are able to obtain broad knowledge covering a wide range of domains \cite{rae_scaling_2021, hoffmann_empirical_2022, touvron_llama_2023, du_glam_2022, guo2023evaluating}, including commonsense knowledge \cite{west-etal-2022-symbolic, bian2023chatgpt, bang2023multitask}. However, commonsense reasoning is still regarded as a major challenge for LLMs \cite{zhou2020evaluating, bhargava2022commonsense}. Studies disclose that LLMs fall short in performing adequate commonsense reasoning \cite{wei_chain--thought_2022}. For example, ChatGPT\footnote{https://openai.com/blog/chatgpt} does not precisely know what the needed commonsense knowledge for answering a specific question is (e.g., questions in social and temporal domains) \cite{bian2023chatgpt}.

To mitigate this issue, we propose CORECODE (\textbf{Co}mmonsense \textbf{Re}asoning and \textbf{Co}nflict \textbf{De}tection in dialogues), a dataset that contains abundant commonsense knowledge manually annotated on Chinese dyadic dialogues, to assess how much commonsense knowledge the LLMs have gained and how well they can be improved in commonsense reasoning and conflict detection with the annotated knowledge in CORECODE.

Specifically, we focus on annotating fine-grained commonsense knowledge in multi-turn dyadic dialogues. The knowledge annotated in a dialogue is context-sensitive and grounded exclusively in that particular dialogue. Inspired by the annotation convention used in task-oriented dialogue, in which dialogue states are denoted in the form of ``domain: slot = value'', e.g. ``hotel: price range = moderate'' \cite{budzianowski_multiwoz_2018, zhu-etal-2020-crosswoz, quan_risawoz_2020}, we standardize the representation of commonsense knowledge in open-domain dialogues also in the form of ``domain: slot = value''. We categorize commonsense knowledge into three dimensions, namely entity, event, and social interaction, and then construct an ontology over these dimensions, which defines all possible domains for each dimension and all possible slots for each domain. Thanks to the guidance of this ontology, crowdsourcing annotators are able to conveniently annotate fine-grained commonsense knowledge in a consistent way.

Over the curated dataset, we develop six benchmark tasks: commonsense knowledge filling, commonsense knowledge generation, commonsense conflict phrase detection, domain identification, slot identification and event causal inference. These tasks, organized in different forms (e.g., multiple-choice questions, span extraction, text generation), facilitate the evaluation and enhancement of commonsense reasoning in LLMs.

We conduct numerous experiments on CORECODE, attempting to explore two main research questions: (1) Can LLMs master and apply commonsense knowledge well enough to achieve good performance on these tasks? (2) How much further improvements can be obtained by LLMs if they are fine-tuned on CORECODE? Extensive experiments demonstrate that our benchmark tasks are challenging for existing Chinese LLMs, as all evaluated LLMs perform poorly on most tasks. We also show that although the performance of LLMs improves after being fine-tuned on CORECODE, they fail to obtain robust commonsense reasoning ability. When perturbations are introduced, the fine-tuning performance has significantly dropped. 

\section{Related Work}

A variety of datasets and benchmarks focusing on different aspects of commonsense knowledge over textual inputs have been proposed, including science common sense datasets ARC \cite{clark_think_2018} and QASC \cite{khot_qasc_2020}, temporal common sense dataset MC-TACO \cite{zhou-etal-2019-going}, numerical common sense dataset NumerSense \cite{lin-etal-2020-birds}, event common sense dataset HellaSWAG \cite{zellers-etal-2019-hellaswag}, physical common sense dataset PIQA \cite{bisk_piqa_2020}, social common sense dataset Social IQA \cite{sap-etal-2019-social} and general common sense datasets CommonsenseQA \cite{talmor-etal-2019-commonsenseqa}, OpenBookQA \cite{mihaylov-etal-2018-suit}, and WSC \cite{levesque_winograd_2012}. These datasets only examine the model's knowledge and ability in a certain commonsense aspect in the form of multiple-choice questions.

Meanwhile, there have also been many studies devoted to annotating commonsense knowledge involved in utterances in dialogues. ATOMIC \cite{sap_atomic_2019, hwang_comet-_2021} is one such dataset that consists of a large set of inference types. However, ATOMIC is context-insensitive, as its commonsense reasoning operates on phrases taken out of context, disregarding whether an event is performed by the same individual. TIMEDIAL \cite{qin-etal-2021-timedial} focuses on the time reasoning ability of language models in dialogues, while CICERO \cite{ghosal-etal-2022-cicero} provides cause, subsequent events, prerequisites, motivations, and emotional reactions for utterances in dialogues, focusing on these five event-related reasoning types. Both datasets cover only a specific aspect of commonsense knowledge. CIDER \cite{ghosal-etal-2021-cider} extracts knowledge in dialogues into knowledge triplets, which covers fewer commonsense knowledge types than us. For example, \emph{subsequent event}, \emph{subsequent emotional reaction}, \emph{frequency} are beyond the scope of CIDER.

To the best of our knowledge, CORECODE is the first large-scale Chinese dialogue-oriented commonsense knowledge annotation dataset involving comprehensive commonsense knowledge in three dimensions: entity, event, and social interaction, covering a large number of perspectives such as attributes, time, space, and causality. Yet another feature that must be emphasized is that within CORECODE, we manually provide phrases corresponding to the phrases in an original dialogue, which are against common sense in that context. This aims to probe the model's capacity to detect and locate such phrases that are inconsistent with the context in terms of commonsense reasoning. 

\section{Dataset Creation}


The raw data of CORECODE is derived from NaturalConv \cite{wang2021naturalconv} and DuLeMon \cite{xu2022long} datasets, both of which contain multi-turn dialogues between two people. Dialogues in NaturalConv involve a variety of topics (including but not limited to sports, entertainment, and technology). We first take an automatic screening method to identify dialogues that are rich in commonsense knowledge, following \citet{Zhou2021CommonsenseFocusedDF}.

Specifically, we first identify candidate concepts (nouns, verbs, adjectives) in each turn of a dialogue using part-of-speech tagging. We then query the ConceptNet using the identified concepts in each utterance to obtain a list of one-hop commonsense triples of the form $(e_1, r, e_2)$. Next, we check if entity $e_2$ from the triple appears in the concept set of the succeeding utterance. If there is a match, it indicates a potential commonsense link between the two utterances.

Unlike \citet{Zhou2021CommonsenseFocusedDF} who retain dialogues with only one commonsense triple match, we employ a stricter criterion by retaining dialogues where more than three commonsense triple matches are detected. This selection ensures that the kept dialogues possess a substantial amount of commonsense reasoning. The statistics of the screening results on NaturalConv and DuLeMon are shown in Appendix \ref{appendix:data}.

Moreover, to differentiate between the two sides of the conversation, we employ the notation ``A: '' or ``B: '' preceding each utterance to denote the respective speaker. 

\subsection{Data Annotation}

Over the selected dialogues, we perform commonsense knowledge annotation. To guarantee the consistency of annotations across multiple crowd-sourced workers, we adopt a standardized annotation procedure.


We categorize commonsense knowledge in everyday conversations into three dimensions: \emph{entity}, \emph{event}, and \emph{social interaction}. Crowd-sourced workers first need to identify specific instances under these three dimensions from dialogues. Then, with the assistance of linguists, we divide each of these three dimensions into multiple domains to which their commonsense knowledge belongs, and define different slots for each domain, forming a two-level hierarchical taxonomy. Such design is guided by three fundamental principles: \emph{coverage}, \emph{exclusivity}, and \emph{easiness}. The \emph{coverage} rule ensures that the commonsense knowledge system encompasses nearly all conceivable types of commonsense knowledge in dialogues. \emph{Exclusivity} mandates that each commonsense knowledge type remains distinct, devoid of any overlap with other types. Lastly, the \emph{easiness} principle indicates that the commonsense knowledge system is straightforward for annotators to employ. With this convention, crowd-sourced workers are instructed to annotate the identified instances with commonsense knowledge in the form of ``domain: slot = value''. In addition to such annotations, they are also required to provide phrases that, in terms of common sense, conflict with the original textual context. Below, we describe each step in detail.

\subsubsection{Entity, Event, Social Interaction Recognition.}
The first step of the annotation process is to identify specific instances of entity, event, and social interaction that exist in dialogues, according to the following definitions.

\begin{itemize}
    \item \textbf{Entities} refer to objectively existing and distinguishable physical objects in the real world, either representing a general category of people or things, such as ``\emph{cats}'', ``\emph{movies}'', or referring to specific individuals or objects, such as ``\emph{Yao Ming}'', ``\emph{Wolf Warrior}'', etc.
    \item \textbf{Events} are typically text spans in the form of ``subject + predicate'' or ``subject + predicate + object''. They are fine-grained semantic units that describe the state of entities and their actions \cite{zhou-etal-2022-claret}. For example, ``\emph{He looks very excited}'' describes the state of the subject, and ``\emph{He broke his toy}'' illustrates an action where the subject interacts with the object.
    \item \textbf{Social interactions} refer to the set of rules and guidelines that constrain people's behavior when interacting with others. It encompasses a collection of social norms and customs that people are expected to adhere to \cite{bian2023chatgpt}. For instance, ``\emph{It is customary to knock on the door before entering someone else's room}''.
\end{itemize}

\subsubsection{Annotation of Involved Commonsense Knowledge.}
Under each of the three dimensions, we define domains and slots. For entities, we divide the relevant commonsense knowledge into three corresponding domains: attribute, comparison, and space. These domains capture specific properties of the object itself, relationships between the object and other objects, and relationships between the object and the spatial environment in which it is located, respectively. Under each domain, there are further divisions into different slots. For example, under the attribute domain, there are slots ``Is'', ``Is A'', ``Has'', ``Is Made Of'', and so on. For events, the relevant commonsense knowledge includes the prerequisite, cause, and consequence of an event, as well as the temporal and spatial factors associated with the event. For social interactions, we focus on the social norms that humans follow. Instead of subdividing into multiple domains, we divide seven slots under the social norms domain. There are 9 domains and 37 slots included in the three dimensions in total. The full inventory of all domains and slots is listed in Appendix \ref{appendix:slots}.

The second step of the annotation process is to label each entity, event or social interaction instance with its commonsense knowledge in the form of ``domain: slot = value''. The annotated ``value'' does not necessarily need to be an exact span extracted from the original dialogue, but can be a grammatically correct and semantically fluent clause summarized from the dialogue, ensuring that the event and its ``domain: slot = value'' in isolation is informationally complete and logically consistent. It has been emphasized to annotators that for the ``event cause'' slot in the ``cause'' domain and the ``subsequent event'' slot in the ``consequence'' domain, they should be annotated in the form of an event, i.e., in the ``subject + predicate'' form or ``subject + predicate + object'' form.


In addition, the annotators need to indicate which phrases or clauses in the original dialogue led to the identification of this commonsense knowledge, so as to provide a basis for the next step.

\subsubsection{Rewriting of Commonsense Conflict Phrases.}
Finally, for each set of phrases from the original dialogue indicated in the previous step, annotators are required to choose one phrase and provide it with the following two commonsense conflict phrases:

(1) \textbf{Commonsense Conflict Phrase 1}: This phrase should be obtained by conforming to the minimal modification principle, i.e., modifying only one or two words in the original phrase. There should be a commonsense conflict or error after using this phrase to replace the original phrase in the dialogue.

(2) \textbf{Commonsense Conflict Phrase 2}: This phrase should be created by modifying as many words as possible in the original phrase in compliance with the maximum modification principle. When constructing this phrase, annotators can include words that appear in the dialogue to maintain consistency with the dialogue's context. However, it is crucial to ensure as much as possible that the meaning of this phrase differs from the Commonsense Conflict Phrase 1.

The purpose of this annotation step is to explore whether LLMs are able to detect the location of phrases that conflict with the dialogue context in terms of common sense. Therefore, annotators must ensure that after replacing the original phrase in the dialogue with the annotated conflict phrase, there should be only a commonsense error while the dialogue maintains grammatically correct and fluent.

To comprehensively evaluate the commonsense reasoning ability of LLMs, we propose two distinct annotated subsets with varying difficulty levels. During the annotation procedure on 9.7K dialogues, we represent the subject and object of events using the speaker indicators ``A'' or ``B'' from the dialogue and group these annotated instances as an EASY set. A HARD set is annotated on another 10K dialogues, where ``x'' is uniformly employed to denote the subject of all events, while ``y'' is used to represent the predicate of all events, regardless of the dialogue participant to whom the event pertains. Significant challenges in reasoning through events are provided in the HARD set, as LLMs are required to first deduce and locate the event initiator before reasoning.

\subsection{Annotation Quality Control}

In order to standardize the annotation form and control the quality of common sense annotations, we design and develop a knowledge acquisition platform where crowd-sourced workers need to properly click on the appropriate buttons and fill in the corresponding values given the dialogue history.


We adopt a very strict quality control protocol to ensure the quality of annotations. First, we train two reviewers with 200 dialogues. The annotation consistency of the two reviewers is high, with an average Cohen’s Kappa \cite{mchugh2012interrater} of 80.7\% across the annotation tasks. We only hire annotators who have relevant experience in text annotation, e.g., those who have participated in annotation tasks such as Chinese multi-turn dialogue writing and correction, entity extraction or syntactic structure annotation in Chinese texts.

Second, 200 candidate workers participate in a pre-annotation stage. They adhere to the prescribed rules to annotate dialogues. The two reviewers will review annotations of these participants to distinguish whether the annotations meet the requirements. The process has an elimination rate of roughly 80\%, with 43 labelers passing this stage.

Third, we proceed to the training phase. We divided the participants into groups of 5 people each. We train 1-2 quality inspectors within each group, who in turn are responsible for the instruction of the annotators. During this progression, quality inspectors evaluate the rule comprehension and error correction capabilities of the annotators. Those who do not meet the criteria are subjected to further training or eliminated from the process.

At last, 6 quality inspectors with an average Cohen’s Kappa of 59.4\%, as well as 15 annotators, proceed to the formal annotation stage. We take iterative verification and revision during this stage. Any data deemed unsatisfactory will be returned for revision until they are qualified.

\subsection{Overall Statistics}
The overall statistics of the annotated dataset are shown in Table 1. After annotating on 19.7K dialogues, we obtain 76,787 annotations, each consisting of the original dialogue, an entity/event/social interaction instance, a commonsense knowledge represented by a domain-slot-value triplet, the involved phrase from the original dialogue, and two commonsense knowledge conflict phrases. The average number of turns and tokens per dialogue is 19.40 and 501.58, indicating that the dialogues annotated are quite long and informative. Since the knowledge in the social interaction dimension is mainly used to constrain our behavior, but is rarely mentioned in our dialogues, there is little commonsense knowledge annotated for this dimension. The three dimensions of entity, event and social interaction annotations account for 58.42\%, 41.54\% and 0.03\% of overall annotations, respectively.

\begin{table}[htbp]
  \centering
  \resizebox{0.9\columnwidth}{!}{
    \begin{tabular}{lccc}
    \toprule
          & HARD  & EASY  & Total \\
    \toprule
    \# dialogues & 10,000 & 9,700 & 19,700 \\
    Max. turns per dialogue & 26    & 26    & 26 \\
    Min. turns per dialogue & 14    & 16    & 15 \\
    Avg. turns per dialogue & 18.69 & 20.10  & 19.40  \\
    Max. \# tokens per dialogue & 1,002 & 953   & 977.5 \\
    Min. \# tokens per dialogue & 194   & 231   & 212.5 \\
    Avg. \# tokens per dialogue & 464.18 & 538.98 & 501.58 \\
    Avg. \# tokens per turn & 24.83 & 26.81 & 25.82 \\
    \midrule
    \# annotated instances & 37,777 & 39,010 & 76,787 \\
    \# annotated entities & 21,320 & 23,541 & 44,861 \\
    \# annotated events & 16,439 & 15,461 & 31,900 \\
    \# annotated social interactions & 18    & 8     & 26 \\
    \midrule
    \# domain-slot-value triplets & 37,777 & 39,010 & 76,787 \\
    \# commonsense conflict phrases & 75,554 & 78,020 & 153,574 \\
    Avg. \# tokens of conflict phrase 1 & 4.81  & 5.31  & 5.06 \\
    Avg. \# tokens of conflict phrase 2 & 4.74  & 4.91  & 4.83  \\
    \bottomrule
    \end{tabular}%
    }
  \caption{Overall statistics of the CORECODE dataset.}
  \label{tab:stat}%
\end{table}%

\section{Benchmark Tasks}

We use our dataset as a testbed and define 6 tasks in different forms, attempting to evaluate dialogue-level commonsense reasoning capabilities of Chinese LLMs. For each task, we provide both its definition and associated prompt that is constructed to allow LLMs to complete the task in the continuation to the prompt. 

\subsection{Commonsense Knowledge Filling}

\subsubsection{Task definition.} This task is to fill desirable commonsense knowledge into a masked dialogue where a commonsense phrase is replaced with [MASK]. In order to automatically assess the performance of the task, we formulate the task in the form of multiple-choice questions.

\subsubsection{Prompt.} The input prompt to LLMs for this task consists of the question, masked dialogue, answer choices, and suffix: question $\backslash$n masked dialogue $\backslash$n (a) $phrase_1$ (b) $phrase_2$ (c) $phrase_3$ $\backslash$n ``answer: the correct option is''. The three phrases are the corresponding masked commonsense phrase and two manually composed commonsense conflict phrases. See Appendix \ref{appendix:examples} for examples of all tasks.

\subsection{Commonsense Knowledge Generation}

\subsubsection{Task definition.} We frame this task as a generative task that takes the annotated commonsense knowledge values as ground truth and asks LLMs to generate the values according to the dialogue context.

\subsubsection{Prompt.} The input prompt is formatted as: dialogue $\backslash$n question $\backslash$n ``answer:'', where the question is formed by the entity/event/social interaction and its annotated slot through a predefined template and some explanatory text.

\subsection{Commonsense Conflict Phrase Detection}

\subsubsection{Task definition.} We define this task as a span extraction task. We replace the corresponding phrases in the original dialogue with the annotated commonsense conflict phrases, and ask LLMs to extract the commonsense conflict phrases.

\subsubsection{Prompt.} The input prompt is in the form of: dialogue with replaced commonsense conflict phrases $\backslash$n question $\backslash$n ``answer:''.

\subsection{Domain Identification}

\subsubsection{Task definition.} This task is also defined as a multiple-choice question task. Take the entity dimension as an example, LLMs are required to select the domain to which the relationship between an entity and its annotated value belongs, based on the given dialogue context. Since the social interaction dimension includes a single domain, this task is performed on the entity and event dimensions.

\subsubsection{Prompt.} The prompt is formatted like: question $\backslash$n entity or event $\backslash$n annotated value $\backslash$n dialogue $\backslash$n (a) $domain_1$ (b) $domain_2$ $\cdots$ (x) $domain_n$ $\backslash$n ``answer: the correct domain is''.

\subsection{Slot Identification}

\subsubsection{Task definition.} This task is similar to the Domain Identification task, except that this task is chosen from more fine-grained slot options and executed on all the three dimensions.

\subsubsection{Prompt.} The format of the prompt is: question $\backslash$n entity or event or social interaction $\backslash$n annotated value $\backslash$n dialogue $\backslash$n (a) $slot_1$ (b) $slot_2$ $\cdots$ (x) $slot_n$ $\backslash$n ``answer: the correct option is''.

\subsection{Event Causal Inference}

Causal inference is one of the crucial reasoning abilities of human intelligence, which involves establishing the correct cause-and-consequence relationships between events. These relationships are captured in the ``cause: event cause'' slot and the ``consequence: subsequent event'' slot of our taxonomy. We specially design three generative event causal inference tasks that utilize the annotated knowledge involved in these two slots.

\begin{itemize}
    \item \textbf{Subtask 1: Event Cause Inference.} Given the dialogue and event, LLMs are required to generate the cause of the event.
    \item \textbf{Subtask 2: Subsequent Event Inference.} Given the dialogue and event, the consequence of the event is generated by LLMs.
    \item \textbf{Subtask 3: Clipped Subsequent Event Inference.} Given the event and the truncated dialogue where the context succeeding the event is discarded, we require LLMs to generate the consequence of the event.
\end{itemize}


\section{Experiments}

\subsection{Evaluated LLMs}

We evaluated a diverse list of Chinese LLMs that cover a variety of training processes and scales\footnote{All the experiments in the main paper were conducted on the HARD set. Experimental results on the EASY set are available in Appendix \ref{appendix:easy_set}.}: (1) LLMs only being pre-trained on large-scale training corpora, including GLM-10B \cite{du-etal-2022-glm} and BLOOM-7.1B \cite{scao_bloom_2022}, (2) LLMs being both pre-trained and instruction-tuned, including ChatGLM-6B\footnote{https://github.com/THUDM/ChatGLM-6B}, ChatGLM2-6B\footnote{https://github.com/THUDM/ChatGLM2-6B}, MOSS-SFT-16B\footnote{https://huggingface.co/fnlp/moss-moon-003-sft}, Baichuan-7B\footnote{https://github.com/baichuan-inc/baichuan-7B}, BLOOMZ-1.7B, BLOOMZ-7.1B, BLOOMZ-7.1B-MT \cite{Muennighoff-2022-arxiv-Crosslingual}, and BELLE-7B, which is the SFT version based on BLOOMZ-7.1B-MT. We used two variants of BELLE fined-tuned on 200K and 2M instructions separately, i.e., BELLE-7B-0.2M\footnote{https://huggingface.co/BelleGroup/BELLE-7B-0.2M} and BELLE-7B-2M\footnote{https://huggingface.co/BelleGroup/BELLE-7B-2M}. We also evaluated two variants Chinese-Alpaca-Plus-7B and Chinese-Alpaca-Plus-13B of Chinese-Alpaca-Plus \cite{cui2023efficient}. We conducted experiments using the recommended hyperparameter settings for all LLMs. We also evaluated ChatGPT\footnote{https://openai.com/chatgpt} (i.e., GPT-3.5-turbo) from OpenAI as a reference.

Furthermore, to explore the impact of in-context learning (ICL) on model performance, we also carried out experiments on ChatGLM-6B under the few-shot settings, including 1-shot, 3-shot and 5-shot settings.

\subsection{Evaluation Metrics}

For the commonsense knowledge filling, domain identification and slot identification tasks (we refer these three tasks to the selection tasks), we used the accuracy of selecting the correct answer as the evaluation metric. During inference, we have found that even if we explicitly state in the prompt that models should output only the answer option indicator (i.e. a, b, c, etc.), not all models follow this instruction. There is no uniformity in the form of answers generated by each model. Moreover, sometimes models output answers with rationales attached. In order to avoid the underestimation of the model performance due to the varying output formats, we adopted a series of filtering measures to find the correct answer in the output as much as possible. For example, in the case where the ground-truth is ``(a) premise'', the generated answers ``a'', ``A'', ``(a)'', ``(A)'', ``a)'', ``A)'', ``(a)premise'', ``(a) premise'', ``premise'' are all counted as correct. We conducted a quantitative analysis experiment to showcase the effectiveness of the filtering measures. Detailed results of this experiment can be found in Appendix \ref{appendix:auto_analysis}.

For the span extraction task, i.e., the commonsense conflict phrase detection task, we used F1 and exact match (EM) scores calculated by comparing model outputs to ground-truth answers.

For the two generation tasks, namely the commonsense knowledge generation task and the event causal inference task, we evaluated LLMs with F1 and EM scores together with reference based metrics: BLEU \cite{papineni-etal-2002-bleu}, METEOR \cite{banerjee-lavie-2005-meteor}, ROUGE \cite{lin-2004-rouge} and CIDEr \cite{Vedantam_2015_cider}.

\begin{table*}[htbp]
  \centering
    \resizebox{0.95\textwidth}{!}{
    \begin{tabular}{l|cccccccr|rc}
    \toprule
    \multirow{2}[4]{*}{Model} & \multicolumn{8}{c|}{Commonsense Knowledge Generation}                      & \multicolumn{2}{c}{Commonsense Conflict Phrase Detection} \\
\cmidrule{2-11}          & F1    & \multicolumn{1}{l}{EM} & \multicolumn{1}{l}{BLEU1} & \multicolumn{1}{l}{BLEU2} & \multicolumn{1}{l}{METEOR} & \multicolumn{1}{l}{ROUGE-L} & \multicolumn{1}{l}{CIDEr} & \multicolumn{1}{l|}{ChatGPT Score} & \multicolumn{1}{c}{\qquad \ \ \ \ F1 \ \ \ }    & EM \\
    \midrule
    GLM-10B & 0.023  & 0.000  & 0.000  & 0.000  & 0.032  & 0.000  & 0.001  & \multicolumn{1}{c|}{3.190 } & 0.011  & 0.000  \\
    BLOOM-7.1B & 0.071  & 0.000  & 0.017  & 0.000  & 0.115  & 0.004  & 0.017  & \multicolumn{1}{c|}{3.455 } & 0.024  & 0.000  \\
    \midrule
    ChatGLM2-6B & 0.160  & 0.004  & 0.001  & 0.000  & 0.145  & 0.001  & 0.002  & \multicolumn{1}{c|}{3.940 } & 0.029  & 0.001  \\
    BELLE-7B-0.2M & 0.090  & 0.019  & 0.015  & 0.000  & 0.105  & 0.010  & 0.041  & \multicolumn{1}{c|}{3.265 } & 0.024  & 0.000  \\
    BELLE-7B-2M & 0.111  & 0.008  & 0.004  & 0.000  & 0.140  & 0.003  & 0.010  & \multicolumn{1}{c|}{3.555 } & 0.007  & 0.000  \\
    BLOOMZ-1.7B & 0.388  & 0.234  & 0.234  & 0.000  & 0.164  & 0.234  & 0.585  & \multicolumn{1}{c|}{4.060 } & 0.004  & 0.000  \\
    BLOOMZ-7.1B & 0.438  & 0.284  & 0.282  & 0.000  & 0.199  & 0.283  & 0.707  & \multicolumn{1}{c|}{3.980 } & 0.041  & 0.003  \\
    BLOOMZ-7.1B-MT & 0.435  & 0.300  & 0.300  & 0.000  & 0.184  & 0.300  & 0.750  & \multicolumn{1}{c|}{4.030 } & 0.047  & 0.010  \\
    MOSS-SFT-16B & 0.199  & 0.071  & 0.066  & 0.000  & 0.147  & 0.049  & 0.174  & \multicolumn{1}{c|}{3.780 } & 0.038  & 0.001  \\
    Baichuan-7B & 0.071  & 0.000  & 0.000  & 0.000  & 0.072  & 0.000  & 0.001  & \multicolumn{1}{c|}{3.445 } & 0.002  & 0.000  \\
    Chinese-Alpaca-Plus-7B & 0.129  & 0.015  & 0.014  & 0.000  & 0.099  & 0.015  & 0.039  & \multicolumn{1}{c|}{3.375 } & 0.021  & 0.000  \\
    Chinese-Alpaca-Plus-13B & 0.133  & 0.021  & 0.018  & 0.000  & 0.104  & 0.020  & 0.051  & \multicolumn{1}{c|}{3.490 } & 0.021  & 0.000  \\
    \midrule
    ChatGLM-6B & 0.147  & 0.000  & 0.000  & 0.000  & 0.166  & 0.000  & 0.000  & \multicolumn{1}{c|}{3.745 } & 0.044  & 0.001  \\
    ChatGLM-6B  1-shot & 0.202  & 0.061  & 0.035  & 0.000  & 0.154  & 0.048  & 0.120  & \multicolumn{1}{c|}{3.770 } & 0.038  & 0.002  \\
    ChatGLM-6B  3-shot & 0.274  & 0.115  & 0.091  & 0.000  & 0.175  & 0.111  & 0.277  & \multicolumn{1}{c|}{3.885 } & 0.060  & 0.006  \\
    ChatGLM-6B  5-shot & 0.215  & 0.097  & 0.095  & 0.000  & 0.147  & 0.096  & 0.240  & \multicolumn{1}{c|}{3.685 } & 0.052  & 0.007  \\
    \midrule
    ChatGPT & 0.296  & 0.071  & 0.044  & 0.000  & 0.258  & 0.045  & 0.111  &  \multicolumn{1}{c|}{-}  & 0.104  & 0.021  \\
    \bottomrule
    \end{tabular}%
    }
  \caption{Overall performance of evaluated LLMs on the commonsense knowledge generation and commonsense conflict phrase detection task.}
  \label{tab:gen}%
\end{table*}%

\begin{table}[htbp]
  \centering
  \resizebox{0.73\columnwidth}{!}{
    \begin{tabular}{lccc}
    \toprule
    Model & CKF & DI & SI \\
    \midrule
    GLM-10B & 0.157  & 0.060  & 0.051  \\
    BLOOM-7.1B & 0.329  & 0.108  & 0.039  \\
    \midrule
    ChatGLM-6B & 0.788  & 0.246  & 0.113  \\
    ChatGLM2-6B & 0.818  & 0.286  & 0.153  \\
    BELLE-7B-0.2M & 0.392  & 0.208  & 0.212  \\
    BELLE-7B-2M & 0.599  & 0.169  & 0.109  \\
    BLOOMZ-1.7B & 0.709  & 0.248  & 0.044  \\
    BLOOMZ-7.1B & 0.758  & 0.444  & 0.165  \\
    BLOOMZ-7.1B-MT & 0.695  & 0.341  & 0.168  \\
    MOSS-SFT-16B & 0.445  & 0.353  & 0.110  \\
    Baichuan-7B & 0.416  & 0.071  & 0.055  \\
    Chinese-Alpaca-Plus-7B & 0.584  & 0.385  & 0.060  \\
    Chinese-Alpaca-Plus-13B & 0.510  & 0.126  & 0.449  \\
    \midrule
    ChatGPT & 0.896  & 0.275  & 0.084  \\
    \bottomrule
    \end{tabular}%
    }
  \caption{Overall performance of evaluated LLMs on the three selection tasks. CKF: Commonsense Knowledge Filling. DI: Domain Identification. SI: Slot Identification.}
  \label{tab:sel}%
\end{table}%

\subsection{Performance of LLMs without Fine-tuning}

We report the performance of the selection tasks in Table \ref{tab:sel}, and the performance of the span extraction task and generation tasks in Table \ref{tab:gen}. We can see that CoRECODE is a very challenging benchmark for all evaluated LLMs.


From Table \ref{tab:sel}, we observe that models which are instruction-tuned with SFT significantly outperform models being only pre-trained. The best-performing models across the three tasks are ChatGLM2-6B, BLOOMZ-7.1B, and Chinese-Alpaca-Plus-13B, respectively. Notably, on the slot identification task, Chinese-Alpaca-Plus-13B achieves an outstanding and unparalleled score.


On the commonsense knowledge generation task, BLOOMZ family achieves very high scores, as shown in Table \ref{tab:gen}. After checking the outputs of each model, we have found that models like ChatGLM and BELLE usually generate leading sentences or explanatory reasons in their responses, despite our prompt explicitly instructing them not to do so. In contrast, BLOOMZ-1.7B and BLOOMz-7.1B typically generate relatively short phrases as answers, which is consistent with the form of our annotations. They hence achieve higher scores than other evaluated LLMs.

To exclude the effect of answer form and answer length on the performance, we handed over the outputs of evaluated LLMs to ChatGPT for scoring, the average results of which are also reported in Table \ref{tab:gen}. We described the task to ChatGPT and asked it to score the answers according to our pre-defined scoring criteria (see in Appendix \ref{appendix:chatgpt_prompt}). The average scores obtained by these models vary from 3 to 5. According to our criteria, this suggests that the answers generated by LLMs are more likely to be ``answers that fit the context of the dialogue but are not a specific answer to the question'' or ``answers that are semantically inconsistent with the ground-truth answer but are also correct''.

From Tabel \ref{tab:gen} we also observe that the model performance improves under the few-shot settings. However, the performance under the 5-shot setting is worse than that under the 3-shot setting. This might be due to the long length of our dialogues (as shown in Table \ref{tab:stat}, the average number of tokens in a dialogue is 501). The excessive length of model inputs under the 5-shot setting might lead to a decline in performance.

\begin{figure}[t]
\centering
\includegraphics[width=0.98\columnwidth]{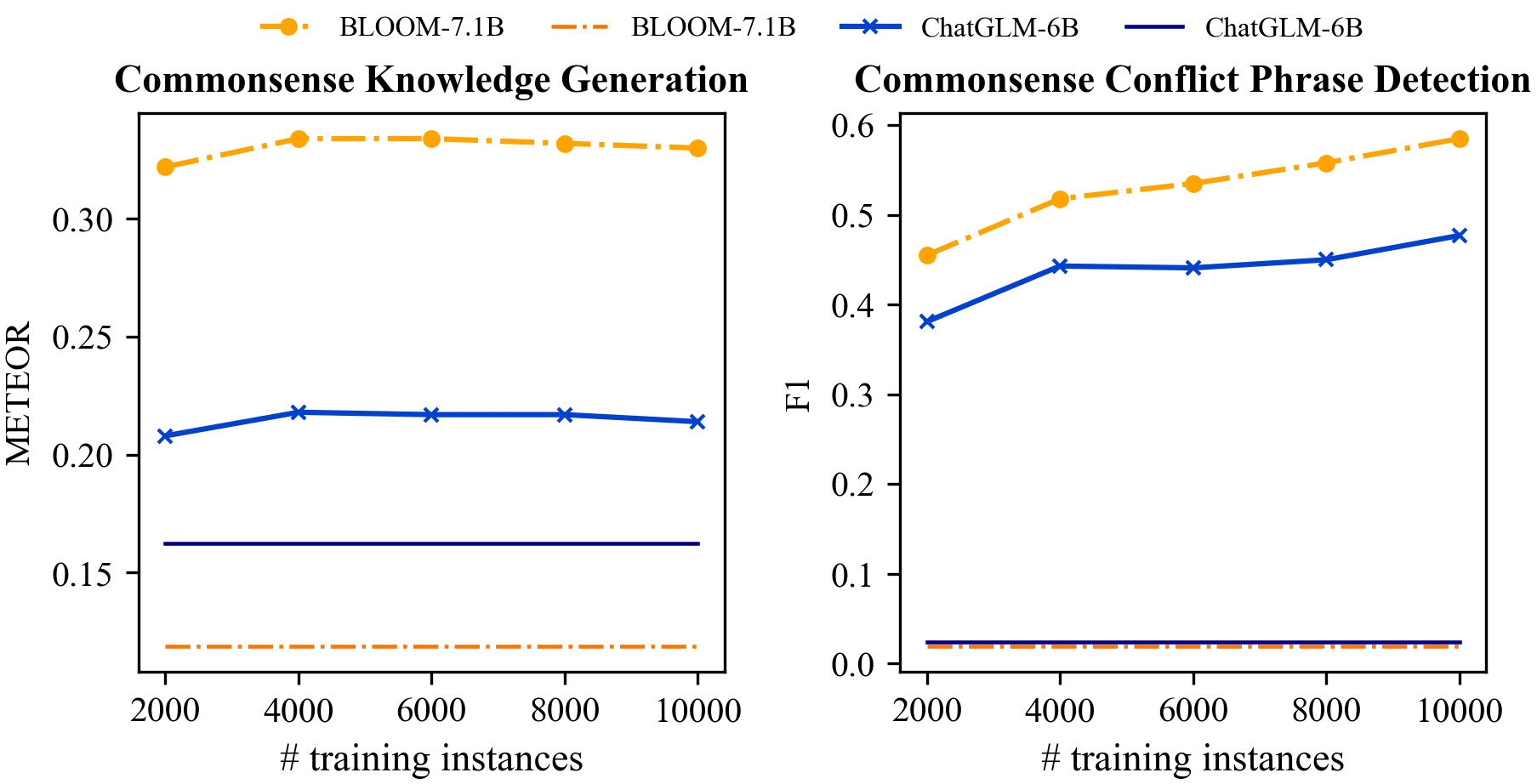}
\caption{Performance of fine-tuned LLMs on the commonsense knowledge generation and commonsense conflict phrase detection task. The horizontal lines show the performance of LLMs without fine-tuning.}
\label{fig:task23}
\end{figure}

\begin{figure*}[t]
\centering
\includegraphics[width=0.83\textwidth]{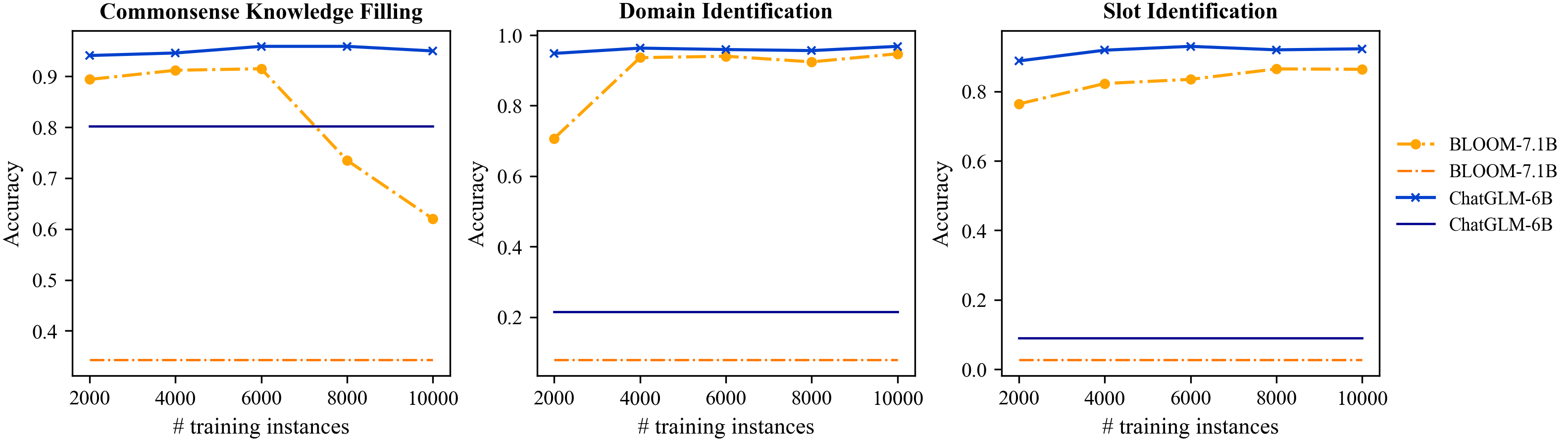} 
\caption{Results of fine-tuned LLMs on the three selection tasks. The horizontal lines show the performance of LLMs without fine-tuning.}
\label{fig:task145}
\end{figure*}

\begin{figure*}[t]
\centering
\hspace{0.63cm}\includegraphics[width=0.88\textwidth]{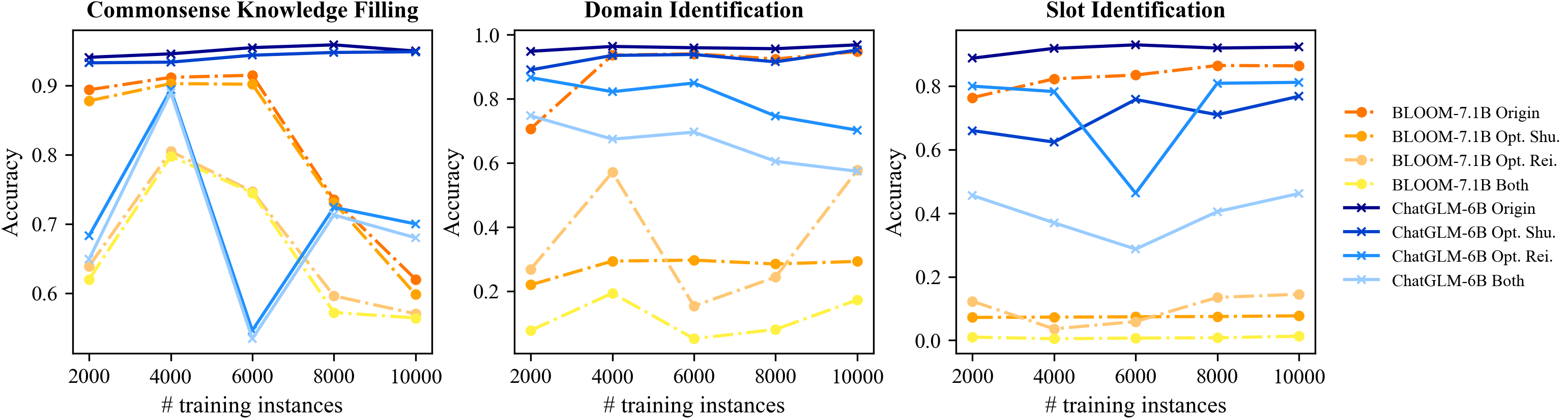}
\caption{Results of fine-tuned LLMs on the perturbed test sets of the three selection tasks, by option re-indicating (Opt. Rei.), option shuffling (Opt. Shu.) and both.}
\label{fig:task145robust}
\end{figure*}

\subsection{Performance of LLMs Being Fine-tuned on CORECODE}

We further evaluated LLMs after they were fine-tuned on CORECODE.

Specifically, we fine-tuned BLOOM-7.1B and ChatGLM-6B on 2K, 4K, 6K, 8K, and 10K examples respectively in the LoRA \cite{DBLP:conf/iclr/HuSWALWWC22} manner, and tested these fine-tuned models on another 2K data. Results on the commonsense knowledge generation and commonsense conflict phrase detection task are shown in Figure \ref{fig:task23}. Fine-tuning on different sizes of data results in large performance gains for both models. On the commonsense conflict phrase detection task, the performance in terms of F1 rises as the size of training data increases. In contrast, on the commonsense knowledge generation task, the performance rises first and then falls as the number of training instances increases, indicating that approximately 4K training instances are sufficient for this task. Training with the same amount of training data for the same duration on both tasks brings more performance gains for BLOOM-7.1B than for ChatGLM-6B. The reason could be that it is easier for BLOOM-7.1B without SFT to acquire such knowledge than ChatGLM-6B with SFT. 

For the three selection tasks, as shown in Figure \ref{fig:task145}, there is a positive correlation between model performance and training data size on most tasks. Both models obtain a substantial improvement after fine-tuning.

\subsection{Robustness Analysis}

Although fine-tuning on CORECODE significantly improves LLMs in commonsense reasoning, is the commonsense reasoning ability that LLMs obtained through fine-tuning robust?

To investigate this question, we conducted three robustness tests on the three selection tasks: (1) option re-indicating, (2) option shuffling, and (3) both. For (1) option re-indicating, we change the option indicators from a, b, c to 1, 2, 3 in the process of forming the prompt. For (2) option shuffling, we shuffle the candidate options and then re-form the input prompt. For (3) both, we implement both option re-indicating and option shuffling.

Experiment results are shown in Figure \ref{fig:task145robust}. We find a decrease in accuracy for both models. Generally, the two LLMs are especially sensitive to option re-indicating, demonstrating larger drops. However, they are more robust to option shuffling, maintaining relatively higher accuracy. The largest performance degradation occurs when both perturbations are executed.

Perturbation causes a dramatic drop to BLOOM-7.1B. When we fine-tune LLMs on CORECODE, we use option indicators, e.g., ``b'', as labels to be learned/predicted. ChatGLM-6B with SFT is better capable of understanding and following instructions than BLOOM-7.1B. It is able to align the indicator to the corresponding answer option during training and combine it with the task instruction to master the involved commonsense reasoning ability. BLOOM-7.1B, however, prefers to learn to answer by memorizing the corresponding input-label mappings. After re-indicating and shuffling answer options, BLOOM-7.1B struggles to answer correctly. For instance, on the slot identification task, our training data has a large number of examples with the label ``b''. BLOOM-7.1B seems to learn such a shortcut incorrectly (i.e., mapping questions to label ``b''). After shuffling answer options (the correct answer indicators are now mostly not ``b''), the model still outputs plenty of ``b'', resulting in very low accuracies, i.e., 0.072, 0.073, 0.074, 0.075, and 0.077 for models trained on 2K, 4K, 6K, 8K, and 10K data, respectively.

\section{Conclusion}
In this paper, we have presented CORECODE, a large-scale commonsense knowledge annotated dialogue dataset with over 76K annotations, and defined 6 benchmark tasks in the form of selection, extraction and generation, to assess the capability of LLMs in learning and applying commonsense knowledge. A diverse list of Chinese LLMs have been evaluated, which achieve poor performance on all tasks, demonstrating the difficulty and utility of the proposed dataset. We have further revealed the robustness issue of LLM commonsense knowledge acquisition via fine-tuning. We hope this work could be used to track and facilitate future advances in context-sensitive LLM commonsense reasoning.

\section*{Acknowledgments}
The present research was supported by Zhejiang Lab (No. 2022KH0AB01). We would like to thank the anonymous reviewers for their insightful comments.

\bibliography{aaai24}

\begin{thebibliography}{53}
\providecommand{\natexlab}[1]{#1}

\bibitem[{Banerjee and Lavie(2005)}]{banerjee-lavie-2005-meteor}
Banerjee, S.; and Lavie, A. 2005.
\newblock {METEOR}: An Automatic Metric for {MT} Evaluation with Improved Correlation with Human Judgments.
\newblock In \emph{Proceedings of the {ACL} Workshop on Intrinsic and Extrinsic Evaluation Measures for Machine Translation and/or Summarization}, 65--72. Ann Arbor, Michigan: Association for Computational Linguistics.

\bibitem[{Bang et~al.(2023)Bang, Cahyawijaya, Lee, Dai, Su, Wilie, Lovenia, Ji, Yu, Chung et~al.}]{bang2023multitask}
Bang, Y.; Cahyawijaya, S.; Lee, N.; Dai, W.; Su, D.; Wilie, B.; Lovenia, H.; Ji, Z.; Yu, T.; Chung, W.; et~al. 2023.
\newblock A multitask, multilingual, multimodal evaluation of chatgpt on reasoning, hallucination, and interactivity.
\newblock \emph{arXiv preprint arXiv:2302.04023}.

\bibitem[{Bauer, Wang, and Bansal(2018)}]{bauer_commonsense_2018}
Bauer, L.; Wang, Y.; and Bansal, M. 2018.
\newblock Commonsense for generative multi-hop question answering tasks.

\bibitem[{Bhargava and Ng(2022)}]{bhargava2022commonsense}
Bhargava, P.; and Ng, V. 2022.
\newblock Commonsense knowledge reasoning and generation with pre-trained language models: A survey.
\newblock In \emph{Proceedings of the AAAI Conference on Artificial Intelligence}, volume~36, 12317--12325.

\bibitem[{Bian et~al.(2023)Bian, Han, Sun, Lin, Lu, and He}]{bian2023chatgpt}
Bian, N.; Han, X.; Sun, L.; Lin, H.; Lu, Y.; and He, B. 2023.
\newblock Chatgpt is a knowledgeable but inexperienced solver: An investigation of commonsense problem in large language models.
\newblock \emph{arXiv preprint arXiv:2303.16421}.

\bibitem[{Bisk et~al.(2020)Bisk, Zellers, Gao, and Choi}]{bisk_piqa_2020}
Bisk, Y.; Zellers, R.; Gao, J.; and Choi, Y. 2020.
\newblock Piqa: Reasoning about physical commonsense in natural language.
\newblock In \emph{Proceedings of the {AAAI} conference on artificial intelligence}, volume~34, 7432--7439.
\newblock ISBN 2374-3468.
\newblock Issue: 05.

\bibitem[{Budzianowski et~al.(2018)Budzianowski, Wen, Tseng, Casanueva, Ultes, Ramadan, and Gašić}]{budzianowski_multiwoz_2018}
Budzianowski, P.; Wen, T.-H.; Tseng, B.-H.; Casanueva, I.; Ultes, S.; Ramadan, O.; and Gašić, M. 2018.
\newblock {MultiWOZ} - A Large-Scale Multi-Domain Wizard-of-Oz Dataset for Task-Oriented Dialogue Modelling.
\newblock In \emph{Proceedings of the 2018 Conference on Empirical Methods in Natural Language Processing}, 5016--5026. Association for Computational Linguistics.
\newblock {EMNLP} 2018.

\bibitem[{Cambria et~al.(2011)Cambria, Song, Wang, and Hussain}]{Cambria2011IsanetteAC}
Cambria, E.; Song, Y.; Wang, H.; and Hussain, A. 2011.
\newblock Isanette: A Common and Common Sense Knowledge Base for Opinion Mining.
\newblock \emph{2011 IEEE 11th International Conference on Data Mining Workshops}, 315--322.

\bibitem[{Clark et~al.(2018)Clark, Cowhey, Etzioni, Khot, Sabharwal, Schoenick, and Tafjord}]{clark_think_2018}
Clark, P.; Cowhey, I.; Etzioni, O.; Khot, T.; Sabharwal, A.; Schoenick, C.; and Tafjord, O. 2018.
\newblock Think you have solved question answering? try arc, the ai2 reasoning challenge.

\bibitem[{Cui, Yang, and Yao(2023)}]{cui2023efficient}
Cui, Y.; Yang, Z.; and Yao, X. 2023.
\newblock Efficient and effective text encoding for chinese llama and alpaca.
\newblock \emph{arXiv preprint arXiv:2304.08177}.

\bibitem[{Du et~al.(2022{\natexlab{a}})Du, Huang, Dai, Tong, Lepikhin, Xu, Krikun, Zhou, Yu, and Firat}]{du_glam_2022}
Du, N.; Huang, Y.; Dai, A.~M.; Tong, S.; Lepikhin, D.; Xu, Y.; Krikun, M.; Zhou, Y.; Yu, A.~W.; and Firat, O. 2022{\natexlab{a}}.
\newblock Glam: Efficient scaling of language models with mixture-of-experts.
\newblock In \emph{International Conference on Machine Learning}, 5547--5569. {PMLR}.
\newblock ISBN 2640-3498.

\bibitem[{Du et~al.(2022{\natexlab{b}})Du, Qian, Liu, Ding, Qiu, Yang, and Tang}]{du-etal-2022-glm}
Du, Z.; Qian, Y.; Liu, X.; Ding, M.; Qiu, J.; Yang, Z.; and Tang, J. 2022{\natexlab{b}}.
\newblock {GLM}: General Language Model Pretraining with Autoregressive Blank Infilling.
\newblock In \emph{Proceedings of the 60th Annual Meeting of the Association for Computational Linguistics (Volume 1: Long Papers)}, 320--335. Dublin, Ireland: Association for Computational Linguistics.

\bibitem[{Ghosal et~al.(2021)Ghosal, Hong, Shen, Majumder, Mihalcea, and Poria}]{ghosal-etal-2021-cider}
Ghosal, D.; Hong, P.; Shen, S.; Majumder, N.; Mihalcea, R.; and Poria, S. 2021.
\newblock {CIDER}: Commonsense Inference for Dialogue Explanation and Reasoning.
\newblock In \emph{Proceedings of the 22nd Annual Meeting of the Special Interest Group on Discourse and Dialogue}, 301--313. Singapore and Online: Association for Computational Linguistics.

\bibitem[{Ghosal et~al.(2022)Ghosal, Shen, Majumder, Mihalcea, and Poria}]{ghosal-etal-2022-cicero}
Ghosal, D.; Shen, S.; Majumder, N.; Mihalcea, R.; and Poria, S. 2022.
\newblock {CICERO}: A Dataset for Contextualized Commonsense Inference in Dialogues.
\newblock In \emph{Proceedings of the 60th Annual Meeting of the Association for Computational Linguistics (Volume 1: Long Papers)}, 5010--5028. Dublin, Ireland: Association for Computational Linguistics.

\bibitem[{Guo et~al.(2023)Guo, Jin, Liu, Huang, Shi, Yu, Liu, Li, Xiong, Xiong et~al.}]{guo2023evaluating}
Guo, Z.; Jin, R.; Liu, C.; Huang, Y.; Shi, D.; Yu, L.; Liu, Y.; Li, J.; Xiong, B.; Xiong, D.; et~al. 2023.
\newblock Evaluating large language models: A comprehensive survey.
\newblock \emph{arXiv preprint arXiv:2310.19736}.

\bibitem[{Hoffmann et~al.(2022)Hoffmann, Borgeaud, Mensch, Buchatskaya, Cai, Rutherford, de~Las~Casas, Hendricks, Welbl, and Clark}]{hoffmann_empirical_2022}
Hoffmann, J.; Borgeaud, S.; Mensch, A.; Buchatskaya, E.; Cai, T.; Rutherford, E.; de~Las~Casas, D.; Hendricks, L.~A.; Welbl, J.; and Clark, A. 2022.
\newblock An empirical analysis of compute-optimal large language model training.
\newblock 35: 30016--30030.

\bibitem[{Hu et~al.(2022)Hu, Shen, Wallis, Allen{-}Zhu, Li, Wang, Wang, and Chen}]{DBLP:conf/iclr/HuSWALWWC22}
Hu, E.~J.; Shen, Y.; Wallis, P.; Allen{-}Zhu, Z.; Li, Y.; Wang, S.; Wang, L.; and Chen, W. 2022.
\newblock LoRA: Low-Rank Adaptation of Large Language Models.
\newblock In \emph{The Tenth International Conference on Learning Representations, {ICLR} 2022, Virtual Event, April 25-29, 2022}. OpenReview.net.

\bibitem[{Hwang et~al.(2021)Hwang, Bhagavatula, Le~Bras, Da, Sakaguchi, Bosselut, and Choi}]{hwang_comet-_2021}
Hwang, J.~D.; Bhagavatula, C.; Le~Bras, R.; Da, J.; Sakaguchi, K.; Bosselut, A.; and Choi, Y. 2021.
\newblock (Comet-) atomic 2020: on symbolic and neural commonsense knowledge graphs.
\newblock In \emph{Proceedings of the {AAAI} Conference on Artificial Intelligence}, volume~35, 6384--6392.
\newblock ISBN 2374-3468.
\newblock Issue: 7.

\bibitem[{Jiang et~al.(2021)Jiang, Bosselut, Bhagavatula, and Choi}]{jiang_im_2021}
Jiang, L.; Bosselut, A.; Bhagavatula, C.; and Choi, Y. 2021.
\newblock “I'm Not Mad”: Commonsense Implications of Negation and Contradiction.
\newblock In \emph{Proceedings of the 2021 Conference of the North American Chapter of the Association for Computational Linguistics: Human Language Technologies}, 4380--4397. Association for Computational Linguistics.

\bibitem[{Khot et~al.(2020)Khot, Clark, Guerquin, Jansen, and Sabharwal}]{khot_qasc_2020}
Khot, T.; Clark, P.; Guerquin, M.; Jansen, P.; and Sabharwal, A. 2020.
\newblock Qasc: A dataset for question answering via sentence composition.
\newblock In \emph{Proceedings of the {AAAI} Conference on Artificial Intelligence}, volume~34, 8082--8090.
\newblock ISBN 2374-3468.
\newblock Issue: 05.

\bibitem[{Levesque, Davis, and Morgenstern(2012)}]{levesque_winograd_2012}
Levesque, H.; Davis, E.; and Morgenstern, L. 2012.
\newblock The winograd schema challenge.
\newblock In \emph{Thirteenth international conference on the principles of knowledge representation and reasoning}.

\bibitem[{Lin et~al.(2019)Lin, Chen, Chen, and Ren}]{lin-etal-2019-kagnet}
Lin, B.~Y.; Chen, X.; Chen, J.; and Ren, X. 2019.
\newblock {K}ag{N}et: Knowledge-Aware Graph Networks for Commonsense Reasoning.
\newblock In \emph{Proceedings of the 2019 Conference on Empirical Methods in Natural Language Processing and the 9th International Joint Conference on Natural Language Processing (EMNLP-IJCNLP)}, 2829--2839. Hong Kong, China: Association for Computational Linguistics.

\bibitem[{Lin et~al.(2020)Lin, Lee, Khanna, and Ren}]{lin-etal-2020-birds}
Lin, B.~Y.; Lee, S.; Khanna, R.; and Ren, X. 2020.
\newblock {B}irds have four legs?! {N}umer{S}ense: {P}robing {N}umerical {C}ommonsense {K}nowledge of {P}re-{T}rained {L}anguage {M}odels.
\newblock In \emph{Proceedings of the 2020 Conference on Empirical Methods in Natural Language Processing (EMNLP)}, 6862--6868. Online: Association for Computational Linguistics.

\bibitem[{Lin(2004)}]{lin-2004-rouge}
Lin, C.-Y. 2004.
\newblock {ROUGE}: A Package for Automatic Evaluation of Summaries.
\newblock In \emph{Text Summarization Branches Out}, 74--81. Barcelona, Spain: Association for Computational Linguistics.

\bibitem[{Liu and Singh(2004)}]{liu2004conceptnet}
Liu, H.; and Singh, P. 2004.
\newblock ConceptNet—a practical commonsense reasoning tool-kit.
\newblock \emph{BT technology journal}, 22(4): 211--226.

\bibitem[{Liu et~al.(2022)Liu, Liu, Lu, Welleck, West, Le~Bras, Choi, and Hajishirzi}]{liu_generated_2022}
Liu, J.; Liu, A.; Lu, X.; Welleck, S.; West, P.; Le~Bras, R.; Choi, Y.; and Hajishirzi, H. 2022.
\newblock Generated Knowledge Prompting for Commonsense Reasoning.
\newblock In \emph{Proceedings of the 60th Annual Meeting of the Association for Computational Linguistics (Volume 1: Long Papers)}, 3154--3169. Association for Computational Linguistics.

\bibitem[{Liu et~al.(2021)Liu, Wan, He, Peng, and Philip}]{liu_kg-bart_2021}
Liu, Y.; Wan, Y.; He, L.; Peng, H.; and Philip, S.~Y. 2021.
\newblock Kg-bart: Knowledge graph-augmented bart for generative commonsense reasoning.
\newblock In \emph{Proceedings of the {AAAI} Conference on Artificial Intelligence}, volume~35, 6418--6425.
\newblock ISBN 2374-3468.
\newblock Issue: 7.

\bibitem[{Lv et~al.(2020)Lv, Guo, Xu, Tang, Duan, Gong, Shou, Jiang, Cao, and Hu}]{lv_graph-based_2020}
Lv, S.; Guo, D.; Xu, J.; Tang, D.; Duan, N.; Gong, M.; Shou, L.; Jiang, D.; Cao, G.; and Hu, S. 2020.
\newblock Graph-Based Reasoning over Heterogeneous External Knowledge for Commonsense Question Answering.
\newblock 34(5): 8449--8456.

\bibitem[{McHugh(2012)}]{mchugh2012interrater}
McHugh, M.~L. 2012.
\newblock Interrater reliability: the kappa statistic.
\newblock \emph{Biochemia medica}, 22(3): 276--282.

\bibitem[{Mihaylov et~al.(2018)Mihaylov, Clark, Khot, and Sabharwal}]{mihaylov-etal-2018-suit}
Mihaylov, T.; Clark, P.; Khot, T.; and Sabharwal, A. 2018.
\newblock Can a Suit of Armor Conduct Electricity? A New Dataset for Open Book Question Answering.
\newblock In \emph{Proceedings of the 2018 Conference on Empirical Methods in Natural Language Processing}, 2381--2391. Brussels, Belgium: Association for Computational Linguistics.

\bibitem[{Muennighoff et~al.(2022)Muennighoff, Wang, Sutawika, Roberts, Biderman, Scao, Bari, Shen, Yong, Schoelkopf, Tang, Radev, Aji, Almubarak, Albanie, Alyafeai, Webson, Raff, and Raffel}]{Muennighoff-2022-arxiv-Crosslingual}
Muennighoff, N.; Wang, T.; Sutawika, L.; Roberts, A.; Biderman, S.; Scao, T.~L.; Bari, M.~S.; Shen, S.; Yong, Z.~X.; Schoelkopf, H.; Tang, X.; Radev, D.; Aji, A.~F.; Almubarak, K.; Albanie, S.; Alyafeai, Z.; Webson, A.; Raff, E.; and Raffel, C. 2022.
\newblock Crosslingual Generalization through Multitask Finetuning.
\newblock \emph{CoRR}, abs/2211.01786.

\bibitem[{Papineni et~al.(2002)Papineni, Roukos, Ward, and Zhu}]{papineni-etal-2002-bleu}
Papineni, K.; Roukos, S.; Ward, T.; and Zhu, W.-J. 2002.
\newblock {B}leu: a Method for Automatic Evaluation of Machine Translation.
\newblock In \emph{Proceedings of the 40th Annual Meeting of the Association for Computational Linguistics}, 311--318. Philadelphia, Pennsylvania, USA: Association for Computational Linguistics.

\bibitem[{Qin et~al.(2021)Qin, Gupta, Upadhyay, He, Choi, and Faruqui}]{qin-etal-2021-timedial}
Qin, L.; Gupta, A.; Upadhyay, S.; He, L.; Choi, Y.; and Faruqui, M. 2021.
\newblock {TIMEDIAL}: Temporal Commonsense Reasoning in Dialog.
\newblock In \emph{Proceedings of the 59th Annual Meeting of the Association for Computational Linguistics and the 11th International Joint Conference on Natural Language Processing (Volume 1: Long Papers)}, 7066--7076. Online: Association for Computational Linguistics.

\bibitem[{Quan et~al.(2020)Quan, Zhang, Cao, Li, and Xiong}]{quan_risawoz_2020}
Quan, J.; Zhang, S.; Cao, Q.; Li, Z.; and Xiong, D. 2020.
\newblock {RiSAWOZ}: A Large-Scale Multi-Domain Wizard-of-Oz Dataset with Rich Semantic Annotations for Task-Oriented Dialogue Modeling.
\newblock In \emph{Proceedings of the 2020 Conference on Empirical Methods in Natural Language Processing ({EMNLP})}, 930--940. Association for Computational Linguistics.

\bibitem[{Rae et~al.(2021)Rae, Borgeaud, Cai, Millican, Hoffmann, Song, Aslanides, Henderson, Ring, and Young}]{rae_scaling_2021}
Rae, J.~W.; Borgeaud, S.; Cai, T.; Millican, K.; Hoffmann, J.; Song, F.; Aslanides, J.; Henderson, S.; Ring, R.; and Young, S. 2021.
\newblock Scaling language models: Methods, analysis \& insights from training gopher.

\bibitem[{Sap et~al.(2019{\natexlab{a}})Sap, Le~Bras, Allaway, Bhagavatula, Lourie, Rashkin, Roof, Smith, and Choi}]{sap_atomic_2019}
Sap, M.; Le~Bras, R.; Allaway, E.; Bhagavatula, C.; Lourie, N.; Rashkin, H.; Roof, B.; Smith, N.~A.; and Choi, Y. 2019{\natexlab{a}}.
\newblock Atomic: An atlas of machine commonsense for if-then reasoning.
\newblock In \emph{Proceedings of the {AAAI} conference on artificial intelligence}, volume~33, 3027--3035.
\newblock ISBN 2374-3468.
\newblock Issue: 01.

\bibitem[{Sap et~al.(2019{\natexlab{b}})Sap, Rashkin, Chen, Le~Bras, and Choi}]{sap-etal-2019-social}
Sap, M.; Rashkin, H.; Chen, D.; Le~Bras, R.; and Choi, Y. 2019{\natexlab{b}}.
\newblock Social {IQ}a: Commonsense Reasoning about Social Interactions.
\newblock In \emph{Proceedings of the 2019 Conference on Empirical Methods in Natural Language Processing and the 9th International Joint Conference on Natural Language Processing (EMNLP-IJCNLP)}, 4463--4473. Hong Kong, China: Association for Computational Linguistics.

\bibitem[{Scao et~al.(2022)Scao, Fan, Akiki, Pavlick, Ilić, Hesslow, Castagné, Luccioni, Yvon, and Gallé}]{scao_bloom_2022}
Scao, T.~L.; Fan, A.; Akiki, C.; Pavlick, E.; Ilić, S.; Hesslow, D.; Castagné, R.; Luccioni, A.~S.; Yvon, F.; and Gallé, M. 2022.
\newblock Bloom: A 176b-parameter open-access multilingual language model.
\newblock \emph{arXiv preprint arXiv:2211.05100}.

\bibitem[{Storks, Gao, and Chai(2019)}]{Storks2019RecentAI}
Storks, S.; Gao, Q.; and Chai, J.~Y. 2019.
\newblock Recent Advances in Natural Language Inference: A Survey of Benchmarks, Resources, and Approaches.
\newblock \emph{arXiv: Computation and Language}.

\bibitem[{Talmor et~al.(2019)Talmor, Herzig, Lourie, and Berant}]{talmor-etal-2019-commonsenseqa}
Talmor, A.; Herzig, J.; Lourie, N.; and Berant, J. 2019.
\newblock {C}ommonsense{QA}: A Question Answering Challenge Targeting Commonsense Knowledge.
\newblock In \emph{Proceedings of the 2019 Conference of the North {A}merican Chapter of the Association for Computational Linguistics: Human Language Technologies, Volume 1 (Long and Short Papers)}, 4149--4158. Minneapolis, Minnesota: Association for Computational Linguistics.

\bibitem[{Touvron et~al.(2023)Touvron, Lavril, Izacard, Martinet, Lachaux, Lacroix, Rozière, Goyal, Hambro, and Azhar}]{touvron_llama_2023}
Touvron, H.; Lavril, T.; Izacard, G.; Martinet, X.; Lachaux, M.-A.; Lacroix, T.; Rozière, B.; Goyal, N.; Hambro, E.; and Azhar, F. 2023.
\newblock Llama: Open and efficient foundation language models.

\bibitem[{Vedantam, Lawrence~Zitnick, and Parikh(2015)}]{Vedantam_2015_cider}
Vedantam, R.; Lawrence~Zitnick, C.; and Parikh, D. 2015.
\newblock CIDEr: Consensus-Based Image Description Evaluation.
\newblock In \emph{Proceedings of the IEEE Conference on Computer Vision and Pattern Recognition (CVPR)}.

\bibitem[{Wang et~al.(2020)Wang, Peng, Ilievski, Szekely, and Ren}]{wang_connecting_2020}
Wang, P.; Peng, N.; Ilievski, F.; Szekely, P.; and Ren, X. 2020.
\newblock Connecting the Dots: A Knowledgeable Path Generator for Commonsense Question Answering.
\newblock In \emph{Findings of the Association for Computational Linguistics: {EMNLP} 2020}, 4129--4140. Association for Computational Linguistics.

\bibitem[{Wang et~al.(2021)Wang, Li, Zhao, and Yu}]{wang2021naturalconv}
Wang, X.; Li, C.; Zhao, J.; and Yu, D. 2021.
\newblock Naturalconv: A chinese dialogue dataset towards multi-turn topic-driven conversation.
\newblock In \emph{Proceedings of the AAAI Conference on Artificial Intelligence}, volume~35, 14006--14014.

\bibitem[{Wei et~al.(2022)Wei, Wang, Schuurmans, Bosma, Xia, Chi, Le, and Zhou}]{wei_chain--thought_2022}
Wei, J.; Wang, X.; Schuurmans, D.; Bosma, M.; Xia, F.; Chi, E.; Le, Q.~V.; and Zhou, D. 2022.
\newblock Chain-of-thought prompting elicits reasoning in large language models.
\newblock 35: 24824--24837.

\bibitem[{West et~al.(2022)West, Bhagavatula, Hessel, Hwang, Jiang, Le~Bras, Lu, Welleck, and Choi}]{west-etal-2022-symbolic}
West, P.; Bhagavatula, C.; Hessel, J.; Hwang, J.; Jiang, L.; Le~Bras, R.; Lu, X.; Welleck, S.; and Choi, Y. 2022.
\newblock Symbolic Knowledge Distillation: from General Language Models to Commonsense Models.
\newblock In \emph{Proceedings of the 2022 Conference of the North American Chapter of the Association for Computational Linguistics: Human Language Technologies}, 4602--4625. Seattle, United States: Association for Computational Linguistics.

\bibitem[{Xu et~al.(2022)Xu, Gou, Wu, Niu, Wu, Wang, and Wang}]{xu2022long}
Xu, X.; Gou, Z.; Wu, W.; Niu, Z.-Y.; Wu, H.; Wang, H.; and Wang, S. 2022.
\newblock Long time no see! open-domain conversation with long-term persona memory.
\newblock \emph{arXiv preprint arXiv:2203.05797}.

\bibitem[{Zellers et~al.(2019)Zellers, Holtzman, Bisk, Farhadi, and Choi}]{zellers-etal-2019-hellaswag}
Zellers, R.; Holtzman, A.; Bisk, Y.; Farhadi, A.; and Choi, Y. 2019.
\newblock {H}ella{S}wag: Can a Machine Really Finish Your Sentence?
\newblock In \emph{Proceedings of the 57th Annual Meeting of the Association for Computational Linguistics}, 4791--4800. Florence, Italy: Association for Computational Linguistics.

\bibitem[{Zhou et~al.(2019)Zhou, Khashabi, Ning, and Roth}]{zhou-etal-2019-going}
Zhou, B.; Khashabi, D.; Ning, Q.; and Roth, D. 2019.
\newblock {``}Going on a vacation{''} takes longer than {``}Going for a walk{''}: A Study of Temporal Commonsense Understanding.
\newblock In \emph{Proceedings of the 2019 Conference on Empirical Methods in Natural Language Processing and the 9th International Joint Conference on Natural Language Processing (EMNLP-IJCNLP)}, 3363--3369. Hong Kong, China: Association for Computational Linguistics.

\bibitem[{Zhou et~al.(2021)Zhou, Gopalakrishnan, Hedayatnia, Kim, Pujara, Ren, Liu, and Hakkani-T{\"u}r}]{Zhou2021CommonsenseFocusedDF}
Zhou, P.; Gopalakrishnan, K.; Hedayatnia, B.; Kim, S.; Pujara, J.; Ren, X.; Liu, Y.; and Hakkani-T{\"u}r, D.~Z. 2021.
\newblock Commonsense-Focused Dialogues for Response Generation: An Empirical Study.
\newblock In \emph{SIGDIAL Conferences}.

\bibitem[{Zhou et~al.(2020)Zhou, Zhang, Cui, and Huang}]{zhou2020evaluating}
Zhou, X.; Zhang, Y.; Cui, L.; and Huang, D. 2020.
\newblock Evaluating commonsense in pre-trained language models.
\newblock In \emph{Proceedings of the AAAI conference on artificial intelligence}, volume~34, 9733--9740.

\bibitem[{Zhou et~al.(2022)Zhou, Shen, Geng, Long, and Jiang}]{zhou-etal-2022-claret}
Zhou, Y.; Shen, T.; Geng, X.; Long, G.; and Jiang, D. 2022.
\newblock {C}lar{ET}: Pre-training a Correlation-Aware Context-To-Event Transformer for Event-Centric Generation and Classification.
\newblock In \emph{Proceedings of the 60th Annual Meeting of the Association for Computational Linguistics (Volume 1: Long Papers)}, 2559--2575. Dublin, Ireland: Association for Computational Linguistics.

\bibitem[{Zhu et~al.(2020)Zhu, Huang, Zhang, Zhu, and Huang}]{zhu-etal-2020-crosswoz}
Zhu, Q.; Huang, K.; Zhang, Z.; Zhu, X.; and Huang, M. 2020.
\newblock {C}ross{WOZ}: A Large-Scale {C}hinese Cross-Domain Task-Oriented Dialogue Dataset.
\newblock \emph{Transactions of the Association for Computational Linguistics}, 8: 281--295.

\end{thebibliography}

\clearpage
\appendix






\section{Data Statistics}\label{appendix:data}
The automated filtering results on NaturalConv and DuLeMon are demonstrated in Table~\ref{tab:preprocess}.

\section{Inventory of All Domains and Slots}\label{appendix:slots}
See Tabel \ref{tab:slots} for the full
inventory of all domains and slots.

\section{Examples of All Tasks}\label{appendix:examples}
\begin{figure*}[t]
\centering
\includegraphics[width=0.98\textwidth]{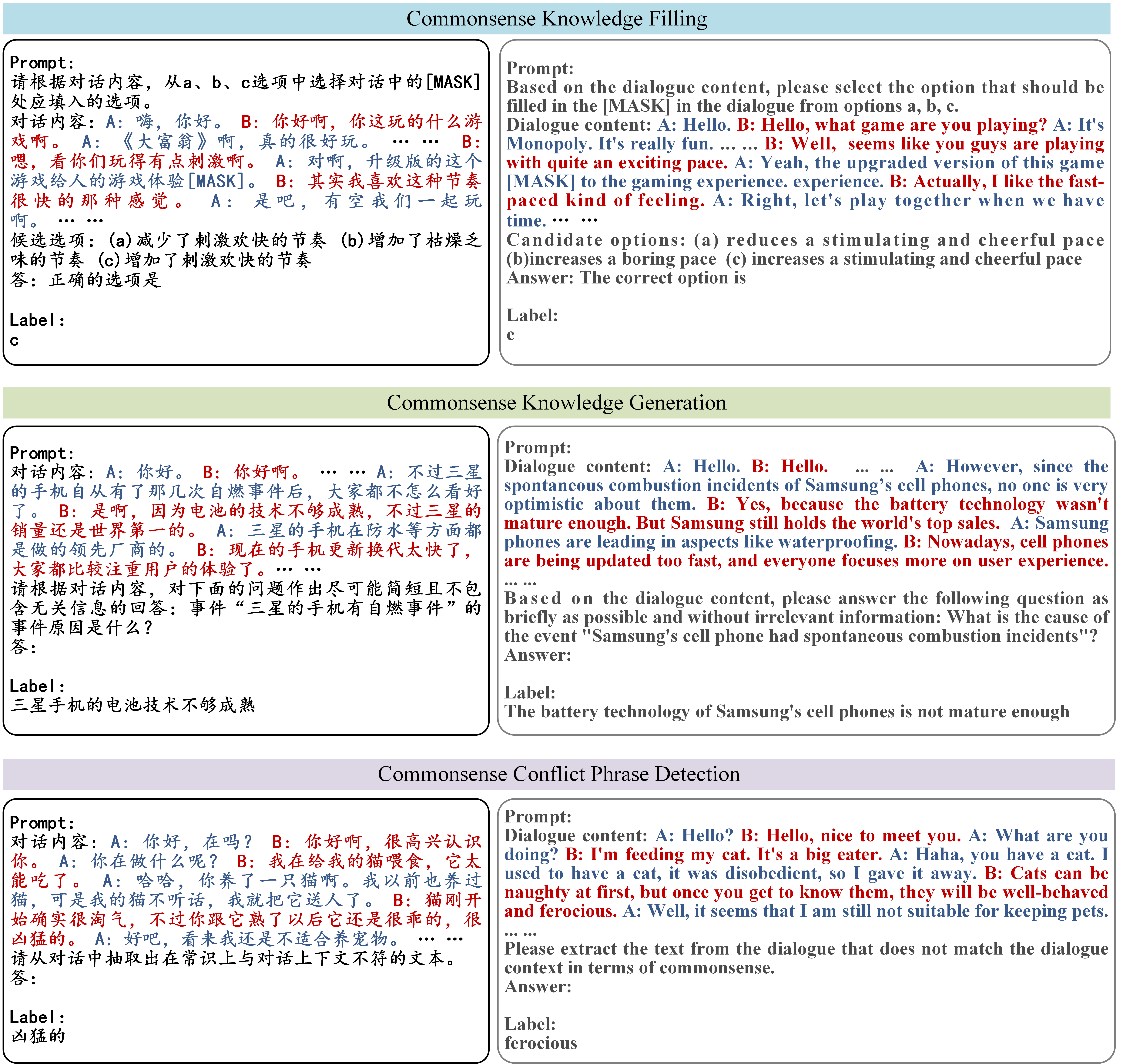} 
\caption{Examples of the commonsense knowledge filling, commonsense knowledge generation and commonsense conflict phrase detection tasks. Due to the extensive length of the dialogues, we only display a relevant section of each dialogue that pertains to the reasoning required for the task, rather than presenting the entire dialogue.}
\label{fig:examples_top3}
\end{figure*}

\begin{figure*}[t]
\centering
\includegraphics[width=0.98\textwidth]{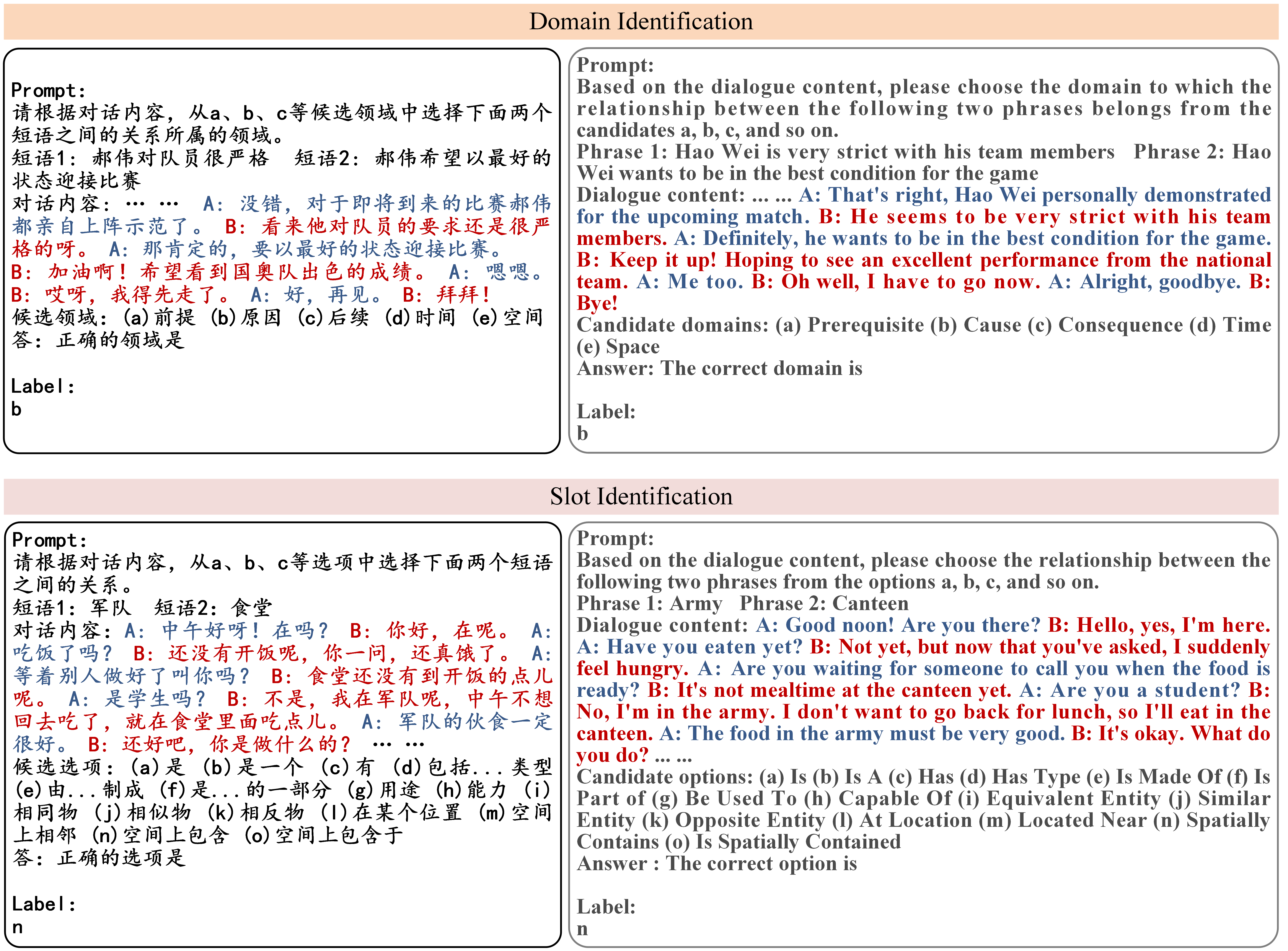} 
\caption{Examples of the domain identification and slot identification tasks. The event causal inference task can be regarded as a special case of the commonsense knowledge generation task, hence no illustrative examples are shown here.}
\label{fig:examples_bottom2}
\end{figure*}

Examples of input prompts for all tasks are shown in Figure \ref{fig:examples_top3} and Figure \ref{fig:examples_bottom2}, where the Chinese text in the black box is the prompt we input to LLMs, and the text in the gray box is the corresponding English translation.

\section{Quantitative Analysis on the Reliability of the Automatic Evaluation}\label{appendix:auto_analysis}
As mentioned in Section 5, we employed an automatic metric, i.e., accuracy, to assess the performance of LLMs on the three selection tasks. Given the highly uncontrollable outputs of LLMs, we applied filtering measures to their outputs to prevent underestimation of LLM performance. To determine the extent to which these filtering measures improve evaluation, we conducted a quantitative analysis on the reliability of the auto evaluation, which is displayed in Table \ref{tab:auto_analysis}. We manually examined 100 outputs from each LLM and recorded the count of correct answers in various scenarios, including: (1) the raw outputs that are counted as correct (``a'', ``(a)'', ``(a)premise'', and ``premise'' are all treated as permissible formats), (2) the filtered outputs that are counted as correct, and (3) the actually correct outputs, which contain outputs that cannot be automatically evaluated as correct due to irregular formats.

The results reveal that the number of answers counted as correct significantly increases when filtering measures are applied, aligning more closely with the actual situation (i.e., the actually correct cases indicated by \# correct). This indicates the effectiveness of the filtering measures, substantially enhancing the evaluation process and mitigating the underestimation of the LLMs' performance.

\section{Prompt for ChatGPT Scoring}\label{appendix:chatgpt_prompt}
\begin{figure*}[t]
\centering
\includegraphics[width=0.96\textwidth]{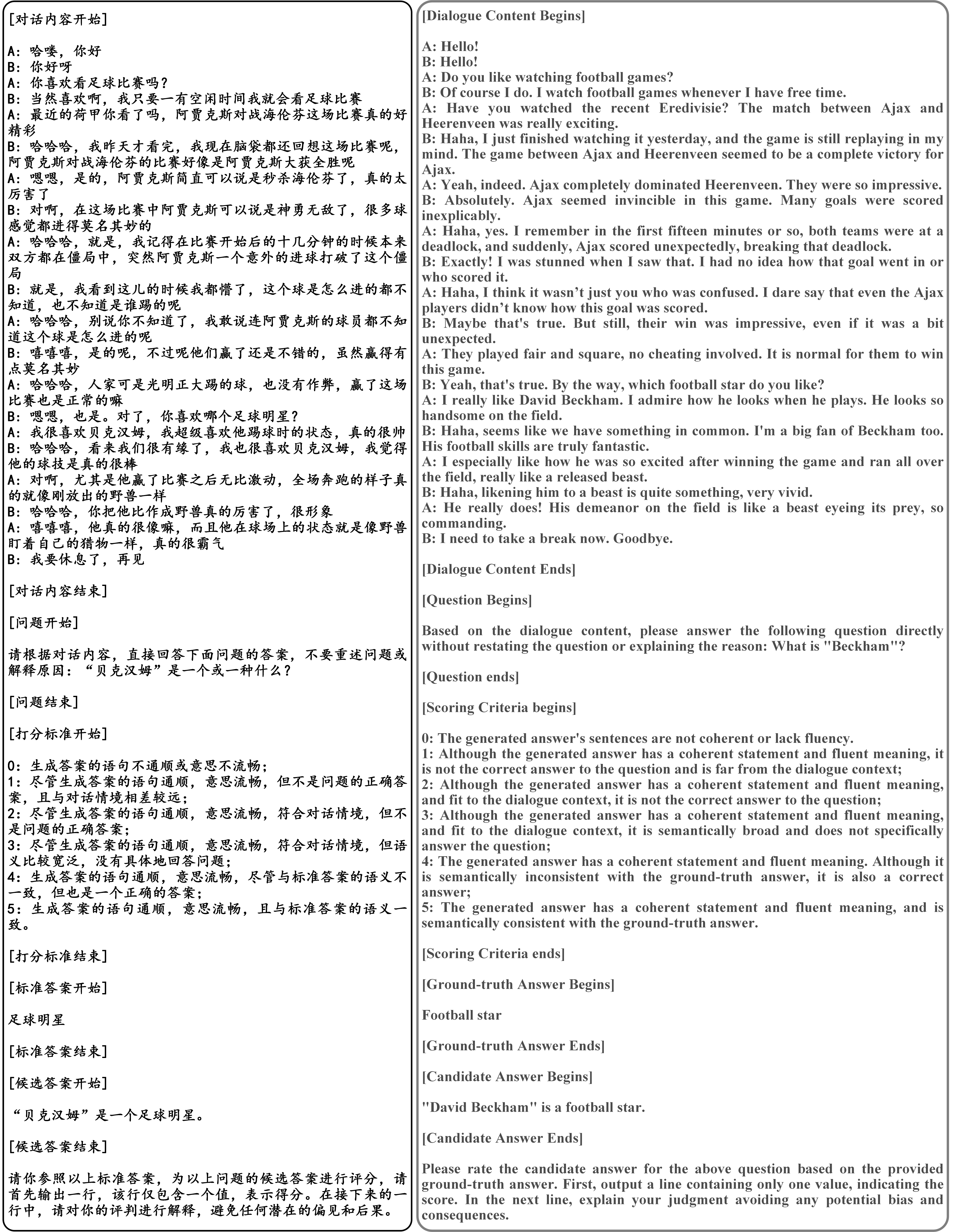} 
\caption{Prompt for ChatGPT to score the outputs of LLMs on the commonsense knowledge generation task. The Chinese text in the black box (left) is the prompt we input to the ChatGPT, while the text in the gray box (right) is the corresponding English translation.}
\label{fig:chatgpt_prompt}
\end{figure*}

Figure \ref{fig:chatgpt_prompt} demonstrates the complete prompt input provided to ChatGPT, which contains detailed scoring criteria, requiring ChatGPT to score the outputs of LLMs on the commonsense knowledge generation task.

\section{Experimental Results on the EASY Set}\label{appendix:easy_set}
As described before, our EASY set and HARD set differ significantly in how subjects and predicates are represented in event annotations. In the EASY set, they are indicated by speaker indicators ``A'' and ``B'', whereas in the HARD set, they are uniformly denoted as ``x'' and ``y''. Consequently, we conducted experiments to assess and compare the performance of the LLMs across the event-related tasks: the commonsense knowledge generation and event causal inference tasks.

\begin{table}[htbp]
  \centering
  \resizebox{0.96\columnwidth}{!}{
    \begin{tabular}{lccc}
    \toprule
    \textbf{Models} & \textbf{\# raw correct} & \textbf{\# filtered correct} & \textbf{\# correct} \\
    \midrule
    GLM-10B & 0     & 18    & 21 \\
    BLOOM-7.1B & 2     & 29    & 32 \\
    \midrule
    ChatGLM-6B & 37    & 85    & 85 \\
    BELLE-7B-2M & 22    & 69    & 69 \\
    BLOOMZ-7.1B & 49    & 76    & 78 \\
    MOSS-SFT-16B & 40    & 40    & 40 \\
    Baichuan-7B & 1     & 40    & 48 \\
    Chinese-Alpaca-Plus-7B & 0     & 62    & 62 \\
    Chinese-Alpaca-Plus-13B & 0     & 53    & 53 \\
    \bottomrule
    \end{tabular}%
    }
  \caption{Results of quantitative analysis of automatical evaluation.}
  \label{tab:auto_analysis}%
\end{table}%

The performance of the commonsense knowledge generation task is reported in Table \ref{tab:easy_set_task2}. We observe that LLMs exhibit slightly better performance on the EASY set compared to the HARD set. This variance can be attributed to the fact that LLMs are required to engage in more intricate reasoning when handling the HARD set. In this scenario, they must first deduce and identify the initiator of the event, followed by subsequent inferences. For instance, when posed with the query \emph{``What is the occurrence time of the event `Person A doesn't go out much?'"}, ChatGLM-6B responds with \emph{``Based on the context of the event, the occurrence time of `Person A doesn't go out much' is `Sunday'"}. Conversely, when confronted with the question \emph{``What is the occurrence time of the event `Person x doesn't go out much?'"}, it gave an erroneous response \emph{``It is not possible to ascertain the occurrence time of the event ``Person x doesn't go out much'"}. It is noteworthy that instances of successful reasoning do exist. For example, when prompted with the question \emph{``What is the cause of the event `Person x is not in a good mood?'"}, ChatGLM-6B adeptly deduced the initiator of the event and subsequently reasoned to deliver the accurate answer: \emph{``The event `Person x is not in a good mood' is caused by the event `Person B has recently lost his job'"}.

The results on the three subtasks of the event causal inference task (namely, the event cause inference task, the subsequent event inference task, and the clipped subsequent event inference task) are demonstrated in Table \ref{tab:easy_set_task6}. On these three subtasks, the results on the HARD set also exhibit some degree of degradation compared to the EASY set. Overall, the notably low performance in these three subtasks underscores the formidable nature of the event causal inference problem for existing Chinese LLMs.


\begin{table*}[htbp]
  \centering
  \resizebox{0.99\textwidth}{!}{
    \begin{tabular}{p{0.05\textwidth}p{0.04\textwidth}p{0.17\textwidth}p{0.43\textwidth}p{0.15\textwidth}p{0.15\textwidth}}
    \toprule
        & \multicolumn{1}{l}{Domain} & Slot  & Description & \multicolumn{2}{l}{Example} \\
    \midrule
    \multicolumn{1}{l}{\multirow{15}[6]{*}{\textbf{Entity}}} & \multicolumn{1}{l}{\multirow{8}[2]{*}{\textbf{ Attribution }}} & \textbf{Is} & States that S presents. Its value is an adjective. & dog   & obedient \\
          &       & \textbf{Is A} & S is a specific instance or a subtype of O. Its value is a noun. & dog   & animal \\
          &       & \textbf{Has} & S contains O. O is part of S. & bird  & wings \\
          &       & \textbf{Has Type} & O is a specific instance or a subtype of S. Contrary to ``Is A''. & movie & romance movie \\
          &       & \textbf{Is Made Of} & S is made of O. & noodles & flour \\
          &       & \textbf{Is Part Of} & S is part of O. O contains S. Contrary to ``Has''. & watermelon rind & watermelon \\
          &       & \textbf{Be Used To} & What S can be used to do. & knife & cut \\
          &       & \textbf{Capable Of} & What S can do. & programmer & coding \\
\cmidrule{2-6}          & \multicolumn{1}{l}{\multirow{3}[2]{*}{\textbf{Comparison}}} & \textbf{Equivalent Entity} & Equivalent entity & movie & film \\
          &       & \textbf{Similar Entity} & Similar entity & pomeranian & bichon frise \\
          &       & \textbf{Opposite Entity} & Opposite entity & spear & shield \\
\cmidrule{2-6}          & \multicolumn{1}{l}{\multirow{4}[2]{*}{\textbf{Space}}} & \textbf{At Location} & O is a typical location for S. & traffic police & road \\
          &       & \textbf{Located Near} & S and O are typically near each other. & school & mall \\
          &       & \textbf{Spatially Contains} & S spatially contains O. & school & canteen \\
          &       & \textbf{Is Spatially Contained} & S is spatially contained by O. & cinema & mall \\
    \midrule
    \multicolumn{1}{l}{\multirow{15}[10]{*}{\textbf{Event}}} & \multicolumn{1}{l}{\textbf{Prerequisite}} & \textbf{Prerequisite} & Prerequisite of the event. & Mary is a good student & Mary usually studies diligently and treats people politely. \\
\cmidrule{2-6}          & \multicolumn{1}{l}{\multirow{4}[2]{*}{\textbf{Cause}}} & \textbf{Event Cause} & An event that directly causes this result. & Mary fell down on her bicycle. & A car rushed out. \\
          &       & \textbf{Emotional Cause} & Motivations, expectations, intentions of the person(s) leading to this result. & Mary fell down on her bicycle. & Mary wanted to be safe. \\
          &       & \textbf{Temporal Cause} & The event occurring as a result of being at a certain time. & Mary fell down on her bicycle. & It's dark. \\
          &       & \textbf{Spatial Cause} & The event occurring as a result of being at a certain spatial location. & Mary fell down on her bicycle. & Mary was close to a car. \\
\cmidrule{2-6}          & \multicolumn{1}{l}{\multirow{4}[2]{*}{\textbf{Consequence}}} & \textbf{Subsequent Event} & Subsequent event resulting from this event. & Mary won the first prize in the competition & Mary's dad rewarded Mary with a new computer. \\
          &       & \textbf{Subsequent Emotional Reaction} & Subsequent emotional reaction resulting from this event. & Mary won the first prize in the competition & Mary's happy. \\
          &       & \textbf{Subsequent Time Change} & Subsequent time change resulting from this event. & Mary has been studying all day. & It's time for dinner. \\
          &       & \textbf{Subsequent Location Change} & Subsequent spatial location change resulting from this event. & Mary has been studying all day. & Mary's back home. \\
\cmidrule{2-6}          & \multicolumn{1}{l}{\multirow{5}[2]{*}{\textbf{Time}}} & \textbf{Occurrence Time} & Time when the event occurrs. & Mary fell down on her bicycle. & last month \\
          &       & \textbf{Start Time} & Time when the event starts. & Mary's sleeping. & 8 p.m. \\
          &       & \textbf{End Time} & Time when the event ends. & Mary's sleeping. & 8 a.m. \\
          &       & \textbf{Duration} & Duration of the event. & Mary's sleeping. & eight hours \\
          &       & \textbf{Frequency} & Frequency of the event. & Mary's sleeping. & every day \\
\cmidrule{2-6}          & \multicolumn{1}{l}{\textbf{Space}} & \textbf{Location} & Location where the event occurrs. & Mary got perfect marks on the exam. & school \\
    \midrule
    \multicolumn{1}{l}{\multirow{7}[2]{*}{\textbf{\makecell{Social \\ interaction}}}} & \multicolumn{1}{l}{\multirow{7}[2]{*}{\textbf{\makecell{Social \\ norms}}}} & \textbf{xAttr} & The attribute of x. & x passed judgment on y. & impolite \\
          &       & \textbf{xIntent} & The intention of x. & x paid for their meal. & Gets on with y. \\
          &       & \textbf{xNeedTo} & What x need to do. & x passed judgment on y. & Apologize to y. \\
          &       & \textbf{xReact} & The reaction of x. & x paid for their meal. & Feels good. \\
          &       & \textbf{xEffect} & The effect on x. & x paid for their meal. & Feels further connected to y \\
          &       & \textbf{yReact} & The reaction of y. & x paid for their meal. & Feels uncomfortable. \\
          &       & \textbf{yEffect} & The effect on y. & x passed judgment on y. & Feels angry. \\
    \bottomrule
    \end{tabular}%
    }
  \caption{Annotated domains and slots in our dataset. S and O denote subject and object, respectively.}
  \label{tab:slots}%
\end{table*}%

\begin{table*}[htbp]
  \centering
  \resizebox{0.99\textwidth}{!}{
    \begin{tabular}{l|ccccccc|ccccccc}
    \toprule
    \multirow{2}[4]{*}{\textbf{Model}} & \multicolumn{7}{c|}{\textbf{Commonsense Knowledge Generation (EASY)}} & \multicolumn{7}{c}{\textbf{Commonsense Knowledge Generation (HARD)}} \\
\cmidrule{2-15}          & \textbf{F1} & \multicolumn{1}{l}{\textbf{EM}} & \multicolumn{1}{l}{\textbf{BLEU1}} & \multicolumn{1}{l}{\textbf{BLEU2}} & \multicolumn{1}{l}{\textbf{METEOR}} & \multicolumn{1}{l}{\textbf{ROUGE-L}} & \multicolumn{1}{l|}{\textbf{CIDEr}} & \textbf{F1} & \multicolumn{1}{l}{\textbf{EM}} & \multicolumn{1}{l}{\textbf{BLEU1}} & \multicolumn{1}{l}{\textbf{BLEU2}} & \multicolumn{1}{l}{\textbf{METEOR}} & \multicolumn{1}{l}{\textbf{ROUGE-L}} & \multicolumn{1}{l}{\textbf{CIDEr}} \\
    \midrule
    GLM-10B & 0.027  & 0.001  & 0.000  & 0.000  & 0.038  & 0.001  & 0.002  & 0.023  & 0.000  & 0.000  & 0.000  & 0.032  & 0.000  & 0.001  \\
    BLOOM-7.1B & 0.083  & 0.000  & 0.018  & 0.000  & 0.131  & 0.003  & 0.018  & 0.071  & 0.000  & 0.017  & 0.000  & 0.115  & 0.004  & 0.017  \\
    \midrule
    ChatGLM-6B & 0.147  & 0.001  & 0.000  & 0.000  & 0.173  & 0.000  & 0.000  & 0.147  & 0.000  & 0.000  & 0.000  & 0.166  & 0.000  & 0.000  \\
    ChatGLM2-6B & 0.166  & 0.006  & 0.001  & 0.000  & 0.181  & 0.001  & 0.001  & 0.160  & 0.004  & 0.001  & 0.000  & 0.145  & 0.001  & 0.002  \\
    BELLE-7B-0.2M & 0.112  & 0.032  & 0.027  & 0.000  & 0.130  & 0.020  & 0.072  & 0.090  & 0.019  & 0.015  & 0.000  & 0.105  & 0.010  & 0.041  \\
    BELLE-7B-2M & 0.110  & 0.019  & 0.008  & 0.000  & 0.140  & 0.006  & 0.021  & 0.111  & 0.008  & 0.004  & 0.000  & 0.140  & 0.003  & 0.010  \\
    BLOOMZ-1.7B & 0.422  & 0.241  & 0.239  & 0.000  & 0.200  & 0.240  & 0.599  & 0.388  & 0.234  & 0.234  & 0.000  & 0.164  & 0.234  & 0.585  \\
    BLOOMZ-7.1B & 0.488  & 0.297  & 0.294  & 0.000  & 0.251  & 0.296  & 0.741  & 0.438  & 0.284  & 0.282  & 0.000  & 0.199  & 0.283  & 0.707  \\
    BLOOMZ-7.1B-MT & 0.474  & 0.306  & 0.305  & 0.000  & 0.221  & 0.306  & 0.764  & 0.435  & 0.300  & 0.300  & 0.000  & 0.184  & 0.300  & 0.750  \\
    MOSS-SFT-16B & 0.208  & 0.074  & 0.058  & 0.000  & 0.178  & 0.110  & 0.208  & 0.199  & 0.071  & 0.066  & 0.000  & 0.147  & 0.049  & 0.174  \\
    Baichuan-7B & 0.075  & 0.002  & 0.001  & 0.000  & 0.069  & 0.002  & 0.004  & 0.071  & 0.000  & 0.000  & 0.000  & 0.072  & 0.000  & 0.001  \\
    Chinese-Alpaca-Plus-7B & 0.126  & 0.018  & 0.016  & 0.000  & 0.107  & 0.018  & 0.045  & 0.129  & 0.015  & 0.014  & 0.000  & 0.099  & 0.015  & 0.039  \\
    Chinese-Alpaca-Plus-13B & 0.126  & 0.022  & 0.019  & 0.000  & 0.105  & 0.020  & 0.051  & 0.133  & 0.021  & 0.018  & 0.000  & 0.104  & 0.020  & 0.051  \\
    \midrule
    Average & 0.197  & 0.078  & 0.076  & 0.000  & 0.148  & 0.079  & 0.194  & 0.184  & 0.074  & 0.073  & 0.000  & 0.129  & 0.071  & 0.183  \\
    \bottomrule
    \end{tabular}%
    }
  \caption{Experimental Results on the commonsense knowledge generation task on the EASY set and HARD set.}
  \label{tab:easy_set_task2}%
\end{table*}%

\begin{table*}[htbp]
  \centering
  \resizebox{0.99\textwidth}{!}{
    \begin{tabular}{l|cccccc|cccccc}
    \toprule
          & \textbf{F1} & \multicolumn{1}{l}{\textbf{EM}} & \multicolumn{1}{l}{\textbf{BLEU1}} & \multicolumn{1}{l}{\textbf{METEOR}} & \multicolumn{1}{l}{\textbf{ROUGE-L}} & \multicolumn{1}{l|}{\textbf{CIDEr}} & \textbf{F1} & \multicolumn{1}{l}{\textbf{EM}} & \multicolumn{1}{l}{\textbf{BLEU1}} & \multicolumn{1}{l}{\textbf{METEOR}} & \multicolumn{1}{l}{\textbf{ROUGE-L}} & \multicolumn{1}{l}{\textbf{CIDEr}} \\
    \midrule
    \textbf{Model} & \multicolumn{6}{c|}{\textbf{Event Cause Inference (EASY)}} & \multicolumn{6}{c}{\textbf{Event Cause Inference (HARD)}} \\
    \midrule
    GLM-10B & 0.039  & 0.000  & 0.000  & 0.064  & 0.000  & 0.000  & 0.033  & 0.000  & 0.000  & 0.057  & 0.000  & 0.000  \\
    BLOOM-7.1B & 0.113  & 0.000  & 0.000  & 0.158  & 0.000  & 0.001  & 0.112  & 0.000  & 0.000  & 0.139  & 0.000  & 0.000  \\
    \midrule
    ChatGLM-6B & 0.142  & 0.000  & 0.000  & 0.222  & 0.000  & 0.000  & 0.141  & 0.000  & 0.000  & 0.214  & 0.000  & 0.000  \\
    ChatGLM2-6B & 0.132  & 0.000  & 0.000  & 0.202  & 0.000  & 0.000  & 0.125  & 0.000  & 0.000  & 0.189  & 0.000  & 0.000  \\
    BELLE-7B-0.2M & 0.101  & 0.000  & 0.000  & 0.149  & 0.000  & 0.000  & 0.101  & 0.001  & 0.000  & 0.140  & 0.000  & 0.000  \\
    BELLE-7B-2M & 0.127  & 0.003  & 0.000  & 0.181  & 0.000  & 0.000  & 0.123  & 0.004  & 0.000  & 0.169  & 0.000  & 0.000  \\
    BLOOMZ-1.7B & 0.244  & 0.020  & 0.017  & 0.145  & 0.017  & 0.042  & 0.214  & 0.013  & 0.011  & 0.124  & 0.011  & 0.028  \\
    BLOOMZ-7.1B & 0.320  & 0.033  & 0.026  & 0.205  & 0.026  & 0.066  & 0.310  & 0.021  & 0.019  & 0.192  & 0.019  & 0.047  \\
    BLOOMZ-7.1B-MT & 0.278  & 0.027  & 0.027  & 0.159  & 0.027  & 0.068  & 0.286  & 0.020  & 0.020  & 0.154  & 0.020  & 0.049  \\
    MOSS-SFT-16B & 0.184  & 0.007  & 0.005  & 0.151  & 0.004  & 0.013  & 0.191  & 0.011  & 0.009  & 0.147  & 0.007  & 0.024  \\
    Baichuan-7B & 0.099  & 0.000  & 0.000  & 0.092  & 0.000  & 0.000  & 0.100  & 0.000  & 0.000  & 0.089  & 0.000  & 0.000  \\
    Chinese-Alpaca-Plus-7B & 0.113  & 0.000  & 0.000  & 0.114  & 0.000  & 0.001  & 0.109  & 0.000  & 0.000  & 0.104  & 0.000  & 0.000  \\
    Chinese-Alpaca-Plus-13B & 0.138  & 0.001  & 0.001  & 0.128  & 0.001  & 0.002  & 0.113  & 0.000  & 0.001  & 0.130  & 0.001  & 0.001  \\
    \midrule
    Average & 0.156  & 0.007  & 0.006  & 0.151  & 0.006  & 0.015  & 0.150  & 0.005  & 0.005  & 0.142  & 0.004  & 0.012  \\
    \midrule
    \textbf{Model} & \multicolumn{6}{c|}{\textbf{Subsequent Event Inference (EASY)}} & \multicolumn{6}{c}{\textbf{Subsequent Event Inference (HARD)}} \\
    \midrule
    GLM-10B & 0.045  & 0.000  & 0.000  & 0.068  & 0.000  & 0.000  & 0.040  & 0.000  & 0.000  & 0.053  & 0.000  & 0.000  \\
    BLOOM-7.1B & 0.121  & 0.000  & 0.000  & 0.164  & 0.000  & 0.000  & 0.106  & 0.000  & 0.001  & 0.148  & 0.000  & 0.002  \\
    \midrule
    ChatGLM-6B & 0.112  & 0.001  & 0.000  & 0.177  & 0.000  & 0.000  & 0.102  & 0.000  & 0.000  & 0.166  & 0.000  & 0.000  \\
    ChatGLM2-6B & 0.142  & 0.002  & 0.000  & 0.186  & 0.000  & 0.000  & 0.118  & 0.000  & 0.000  & 0.159  & 0.000  & 0.000  \\
    BELLE-7B-0.2M & 0.083  & 0.000  & 0.001  & 0.121  & 0.000  & 0.000  & 0.079  & 0.002  & 0.000  & 0.115  & 0.000  & 0.000  \\
    BELLE-7B-2M & 0.117  & 0.004  & 0.001  & 0.150  & 0.001  & 0.002  & 0.108  & 0.005  & 0.000  & 0.136  & 0.000  & 0.000  \\
    BLOOMZ-1.7B & 0.228  & 0.032  & 0.028  & 0.144  & 0.030  & 0.074  & 0.168  & 0.015  & 0.013  & 0.108  & 0.013  & 0.034  \\
    BLOOMZ-7.1B & 0.311  & 0.062  & 0.057  & 0.196  & 0.057  & 0.144  & 0.268  & 0.037  & 0.030  & 0.174  & 0.033  & 0.082  \\
    BLOOMZ-7.1B-MT & 0.288  & 0.050  & 0.049  & 0.174  & 0.049  & 0.122  & 0.233  & 0.023  & 0.023  & 0.142  & 0.023  & 0.058  \\
    MOSS-SFT-16B & 0.164  & 0.021  & 0.016  & 0.124  & 0.012  & 0.041  & 0.132  & 0.005  & 0.004  & 0.106  & 0.003  & 0.009  \\
    Baichuan-7B & 0.072  & 0.000  & 0.000  & 0.085  & 0.000  & 0.000  & 0.065  & 0.000  & 0.000  & 0.074  & 0.000  & 0.000  \\
    Chinese-Alpaca-Plus-7B & 0.107  & 0.001  & 0.001  & 0.106  & 0.001  & 0.002  & 0.092  & 0.001  & 0.001  & 0.093  & 0.001  & 0.002  \\
    Chinese-Alpaca-Plus-13B & 0.123  & 0.002  & 0.002  & 0.109  & 0.002  & 0.004  & 0.109  & 0.000  & 0.000  & 0.099  & 0.000  & 0.001  \\
    \midrule
    Average & 0.147  & 0.013  & 0.012  & 0.139  & 0.012  & 0.030  & 0.125  & 0.007  & 0.005  & 0.121  & 0.006  & 0.014  \\
    \midrule
    \textbf{Model} & \multicolumn{6}{c|}{\textbf{Clipped Subsequent Event Inference (EASY)}} & \multicolumn{6}{c}{\textbf{Clipped Subsequent Event Inference (HARD)}} \\
    \midrule
    GLM-10B & 0.048  & 0.000  & 0.000  & 0.037  & 0.000  & 0.000  & 0.043  & 0.000  & 0.000  & 0.029  & 0.000  & 0.000  \\
    BLOOM-7.1B & 0.089  & 0.000  & 0.000  & 0.115  & 0.000  & 0.000  & 0.084  & 0.000  & 0.000  & 0.108  & 0.000  & 0.000  \\
    ChatGLM-6B & 0.069  & 0.000  & 0.000  & 0.111  & 0.000  & 0.000  & 0.067  & 0.000  & 0.000  & 0.108  & 0.000  & 0.000  \\
    ChatGLM2-6B & 0.083  & 0.000  & 0.000  & 0.113  & 0.000  & 0.000  & 0.073  & 0.000  & 0.000  & 0.104  & 0.000  & 0.000  \\
    BELLE-7B-0.2M & 0.070  & 0.000  & 0.000  & 0.091  & 0.000  & 0.000  & 0.065  & 0.000  & 0.000  & 0.084  & 0.000  & 0.000  \\
    BELLE-7B-2M & 0.045  & 0.000  & 0.000  & 0.054  & 0.000  & 0.000  & 0.056  & 0.000  & 0.000  & 0.062  & 0.000  & 0.000  \\
    BLOOMZ-1.7B & 0.118  & 0.000  & 0.000  & 0.070  & 0.000  & 0.000  & 0.096  & 0.001  & 0.001  & 0.057  & 0.001  & 0.003  \\
    BLOOMZ-7.1B & 0.164  & 0.008  & 0.007  & 0.090  & 0.008  & 0.020  & 0.145  & 0.004  & 0.004  & 0.084  & 0.004  & 0.010  \\
    BLOOMZ-7.1B-MT & 0.133  & 0.004  & 0.004  & 0.069  & 0.004  & 0.011  & 0.116  & 0.002  & 0.002  & 0.063  & 0.002  & 0.006  \\
    MOSS-SFT-16B & 0.096  & 0.002  & 0.001  & 0.077  & 0.001  & 0.004  & 0.092  & 0.001  & 0.001  & 0.072  & 0.001  & 0.003  \\
    Baichuan-7B & 0.059  & 0.000  & 0.000  & 0.074  & 0.000  & 0.000  & 0.052  & 0.000  & 0.000  & 0.063  & 0.000  & 0.000  \\
    Chinese-Alpaca-Plus-7B & 0.069  & 0.000  & 0.000  & 0.064  & 0.000  & 0.000  & 0.067  & 0.000  & 0.000  & 0.057  & 0.000  & 0.000  \\
    Chinese-Alpaca-Plus-13B & 0.055  & 0.000  & 0.000  & 0.048  & 0.000  & 0.000  & 0.061  & 0.000  & 0.000  & 0.051  & 0.000  & 0.000  \\
    \midrule
    Average & 0.084  & 0.001  & 0.001  & 0.078  & 0.001  & 0.003  & 0.078  & 0.001  & 0.001  & 0.072  & 0.001  & 0.002  \\
    \bottomrule
    \end{tabular}%
    }
  \caption{Experimental Results on the three subtasks of the event causal inference task on the EASY set and HARD set.}
  \label{tab:easy_set_task6}%
\end{table*}%

\begin{table*}[htbp]
  \centering
    \begin{tabularx}{0.9\textwidth}{lllllX}
    \toprule
          & \# dialogues & \# 1 triple matches & \# 2 triple matches & \# 3 triple matches & \#  dialogues in CORECODE \\
    \midrule
    NaturalConv & 19919 & 17015 & 12938 & 9145  & 9145 \\
    DuLeMon & 27501 & 24362 & 21926 & 18346 & 10555 \\
    Total & 47420 & 41377 & 34864 & 27491 & 19700 \\
    \bottomrule
    \end{tabularx}%
  \caption{Statistics after screening on NaturalConv and DuLeMon. \# dialogues: total number of dialogues in the dataset. \# 1 triple matches: number of dialogues where more than one commonsense triple matches occurred.}
  \label{tab:preprocess}
\end{table*}%

\end{document}